\pdfoutput=1

\documentclass[11pt]{article}

\usepackage{acl}

\usepackage{times}
\usepackage{latexsym}

\usepackage[T1]{fontenc}

\usepackage[utf8]{inputenc}

\usepackage{hyperref}
\usepackage{url}
\usepackage{graphicx}
\usepackage{booktabs}
\usepackage{amsfonts}
\usepackage[utf8]{inputenc}
\usepackage[ruled]{algorithm2e}
\usepackage{booktabs}
\usepackage{amsmath}
\usepackage{xcolor}
\usepackage{makecell}
\usepackage{tabularx}
\usepackage{microtype}

\newtheorem{theorem}{Theorem}
\newtheorem{remark}{Remark}

\title{Pyramid-BERT: Reducing Complexity via \\ Successive Core-set based Token Selection}

\author{Xin Huang \\
  Amazon AWS, USA \\
  \texttt{xinxh@amazon.com} \\\And
  Ashish Khetan \\
  Amazon AWS, USA \\
  \texttt{khetan@amazon.com} \\\AND
  
  Rene Bidart \\
  University of Waterloo, Canada \\
  \texttt{rbbidart@uwaterloo.ca} \\\And

  Zohar Karnin \\
  Amazon AWS, Israel \\
  \texttt{zkarnin@amazon.com}}

\begin{document}
\maketitle
\begin{abstract}
Transformer-based language models such as BERT~\citep{devlin2018bert} have achieved the state-of-the-art performance on various NLP tasks, but are computationally prohibitive. 
A recent line of works use various heuristics to successively shorten sequence length while transforming tokens through encoders, in tasks such as classification and ranking that require a single token embedding for prediction.
We present a novel solution to this problem, called Pyramid-BERT where we replace previously used heuristics with a {\em core-set} based token selection method justified by theoretical results.
The core-set based token selection technique allows us to avoid expensive pre-training, gives a space-efficient fine tuning, and thus makes it suitable to handle longer sequence lengths. 
We provide extensive experiments establishing advantages of pyramid BERT over several baselines and existing works on the GLUE benchmarks and Long Range Arena \citep{tay2020long} datasets.
\end{abstract}

\section{Introduction}
Transformers \citep{vaswani2017attention} have gradually become a key component for many
state-of-the-art natural language representation models. A recent Transformer based
model BERT \citep{devlin2018bert}, and its variations, achieved the state-of-the-art results on various natural language processing tasks, including machine translation~\citep{wang2019learning,liu2020multilingual}, question-answering~\citep{devlin2018bert,yang2019end}, text classification~\citep{goyal2020power,xu2019bert}, semantic role labeling~\citep{strubell2018linguistically}, 
and so on. However, it takes substantial computational resources to pre-train, fine-tune, or infer such models. 
The complexity of Transformers is mainly 
due to a pipeline of encoders,
each of which contains a multi-head self-attention layer. The self-attention operation scales quadratically with the input sequence length, 
which is a bottleneck especially for long-sequence data.

Given this challenge, intensive efforts have been focused on 
compressing and accelerating Transformers
to reduce the cost of pre-training and fine-tuning.
This work is particularly inspired by the sequence-level NLP tasks such as text classification and ranking.
The state-of-the-art Transformer models for these tasks utilize a single embedding from the top encoder layer, such as the CLS token,
for prediction.
Under such regime, retaining full-length sequence till the last encoder creates unnecessary complexity. We follow a line of works detailed in Section~\ref{sec:related} that aim to gradually reduce the sequence length in the pipeline of encoders. 
On a high level, the existing works have two components: 
\newcommand{\mech}{{\em Select}\ }
\newcommand{\mechtr}{{\em Train-Select}\ }
\mech: a mechanism in charge of reducing the sequence length, either by pruning or pooling, \mechtr: a training or fine-tuning procedure dedicated to this mechanism.

Our main contribution is a novel solution for \mech, motivated from the following observation: As we consider the token representations in top layers, they have increasing redundancy among themselves. We provide a quantitative study demonstrating this in Figure~\ref{fig:cls_similarity}. 
A collection of tokens with high redundancy can intuitively be represented by a small \emph{core-set}, composed of a subset of the tokens. Inspired by this, our solution for \mech is based on the idea of \emph{core-sets}. We provide a theoretically motivated approach that becomes more effective as the redundancy in the representation increases. 
%
This concept separates our work from previous art that provide heuristic techniques for sequence length reduction, or approaches that require expensive training. All of these can in fact be shown to fail in toy examples with high redundancy, e.g.\ the same representation duplicated multiple times.

For \mechtr there is some variety in previous art. Some require a full pre-training procedure, and others require a fine-tuning procedure that works on the full uncompressed model, meaning one that keeps all tokens until the final encoder layer. The high quality of our solution to \mech allows us to simply avoid this additional training phase altogether. The result of this simplification is quite impactful. We obtain a speedup and memory reduction not only to the inference but to the training process itself. This makes it possible to use standard hardware (and training scripts) in the training procedure even for very long sequences.

In Section~\ref{sec: experiments} we provide an empirical comparison of our technique with SOTA alternatives and show that it is superior in eliminating redundancy among the tokens, and thus greatly improves the final classification accuracy. 
In particular, our experiments on the GLUE benchmarks show that our method achieves up to $3$-$3.5$X inference speedup while maintaining an accuracy (mean value over all GLUE datasets) drop of $1.5\%$, whereas the best baseline suffers a $2.5\%$ drop. 
We show that our method can be combined with long-text Transformers such as \textit{Big Bird}~\citep{zaheer2020big} and \textit{Performers}~\citep{choromanski2020rethinking} to further alleviate the quadratic space complexity of the self-attention mechanism. Empirical experiments on the Long Range Arena (LRA)~\citep{tay2020long} show that our model achieves a better trade-off between space complexity reduction and accuracy in comparison to its competitors. In particular, when working with Performers and  reducing the space complexity by 70\%, while baselines suffer a drop in accuracy of 4\% or more, our technique actually improves the accuracy as it acts as a regularizer.

Concluding, our paper provides a novel, theoretically justified technique for sequence length reduction, achieving a speedup and memory reduction for both training and inference of transformers, while suffering significantly less in terms of predictive performance when compared to other existing techniques. Our methods are vetted via thorough  empirical studies comparing it against SOTA methods and examining its different components.

\section{Related Works} \label{sec:related}
There have been a number of interesting attempts, that were aimed at model compression for Transformers, which can be broadly categorized into three directions. The first line of work focus on the redundancy of model parameters. Structure pruning~\citep{mccarley2019pruning,michel2019sixteen,voita2019analyzing,wang2019structured,fan2019reducing,wu2021not}, which removes coherent groups of weights to preserve the original structure of the network, is one common strategy. In addition, various types of distillation techniques~\citep{sanh2019distilbert,sun2019patient,jiao2019tinybert,wang2020minilm} have been proposed to remove encoders by training a compact Transformer to reproduce the output of a larger one. Other strategies include weight quantization~\citep{bhandare2019efficient,zafrir2019q8bert,shen2020q,fan2020training} and weight sharing~\citep{dehghani2018universal,lan2019albert}. 
The second line of work focus on reducing the quadratic operation of the self-attention matrices. The quadratic time and space complexity of the attention mechanism with respect to the input sequence serves the main efficiency bottleneck of Transformers, and thus is prohibitively expensive for training long-sequence data. One popular approach is to sparsify the self-attention operation by restricting each token to only attend a subset of tokens~\citep{child2019generating,kitaev2020reformer,ye2019bp,qiu2019blockwise,ainslie2020etc,zaheer2020big,beltagy2020longformer}. In particular, the most recent \textit{Big Bird}~\citep{zaheer2020big} and \textit{Longformer}~\citep{beltagy2020longformer} introduce sparse models which scale linearly with the input sequence. Another popular approach~\citep{wang2020linformer,choromanski2020rethinking} is to approximate the self-attention to reduce its quadratic complexity to linear, where the most recent \textit{Performers}~\citep{choromanski2020rethinking} provides an unbiased linear estimation of the attention matrices.

The third line, which our work lies in, focus on the redundancy in maintaining a full-length sequence of token-level representation across all encoders~\citep{dai2020funnel,pietruszka2020sparsifying,ye2021tr,kim2020length}. 
This work is particularly inspired by the sequence-level NLP tasks such as text classification and ranking, where the state-of-the-art approach utilizes a single embedding from the top encoder layer of a Transformer, such as the CLS token, for prediction. Under such regime, retaining the full-length sequence till the last encoder creates unnecessary complexity.~\citet{dai2020funnel} applied a simplest strided mean pooling to each sliding window of the sequence to gradually compress tokens. The paper proposed a specialized pre-training procedure for this process that achieves a good balance between speedup and accuracy drop. In comparison we aim to avoid the costly pre-training step and focus on a method that requires only a lightweight fine-tuning.
~\citet{wu2021centroid} define a {\em centroid attention} operator that given centroids, computes their embeddings via an attention mechanism. These centroids are chosen either as a random subset of the original tokens, or via strided mean pooling. This results in a similar technique to that of~\citet{dai2020funnel}, in terms of having a naive sequence length reduction method.
\citet{pietruszka2020sparsifying} provide a variant of length 2 strided mean pooling where instead of taking the unweighted average of each pair, they provide learnable weights via a differentiable linear function. 
In our experiment we did not compare our methods with~\citep{wu2021centroid,pietruszka2020sparsifying} since both worked on a limited (non-standard) collection of datasets and at the time of writing this paper, did not provide code allowing us to reproduce their result. Since our paper is focused on techniques to select a subset of tokens and these papers use either random sampling or pooling, we do not believe a thorough comparison is required.
%
%
\citet{ye2021tr} proposed a reinforcement-learning based technique to rank the tokens, thereby allowing it to remove the least important ones. This RL policy must be trained in an expensive process that requires the full network structure. Since we focus on methods to improve both training and inference, we do not include it in our experiments.
Vision Transformers~\citep{pan2021scalable,heo2021rethinking}, which apply various types of pooling techniques to reduce input length, are specially designed for image data and thus non-trivial to compare with our method.
Early exiting approaches~\citep{xin2020deebert,zhou2020bert}, which allow samples to exit early based on redundancy, are orthogonal to our technique.
~\citet{goyal2020power} developed an attention based mechanism to progressively eliminate tokens in the intermediate encoders in the fine-tuning, while maintaining the classification accuracy. 

\vspace{-0.03in}
\section{Pyramid BERT} \label{sec:our model}
\vspace{-0.03in}
\textbf{Background.} A Transformer model, e.g. BERT, takes a sequence of tokens as input for each input sentence. The sequence of tokens consists of a CLS token followed by the tokens generated by tokenizing the input sentence. 
For batch processing, an appropriate sequence length $N$ is chosen, and shorter input sentences are padded to achieve an uniform length $N$. 
The embedding layer $E$ embeds each token into a vector of real numbers of a fixed dimension. For each input sentence, the token embeddings are transformed through a pipeline of encoders and the self-attention mechanism. 
Each encoder takes all the $N$ embeddings as input, and outputs updated $N$ embeddings of the same dimension. 
The time and space complexity of the self-attention scales quadratically with the input sequence length $N$.

\textbf{Motivation.}  
The state-of-the-art BERT utilizes only the CLS token from the top encoder layer for tasks such as classification and ranking. A natural question is: do we need to propagate all the token embeddings through the entire pipeline of encoders when only the top layer CLS embedding is used for prediction? In general, yes, since the self-attention transforms all the embeddings together, updating each one by capturing information from all the others.
However, if two or more tokens are exact duplicates of each other, ignoring the positional embedding, then one can easily remove the duplicates from the input and modify the self-attention appropriately to get the same CLS embedding at the top layer, and hence the same prediction. This would reduce the number of FLOPs carried out in each self-attention layer. 

\begin{figure}[h]
\begin{center}
\includegraphics[scale=0.23]{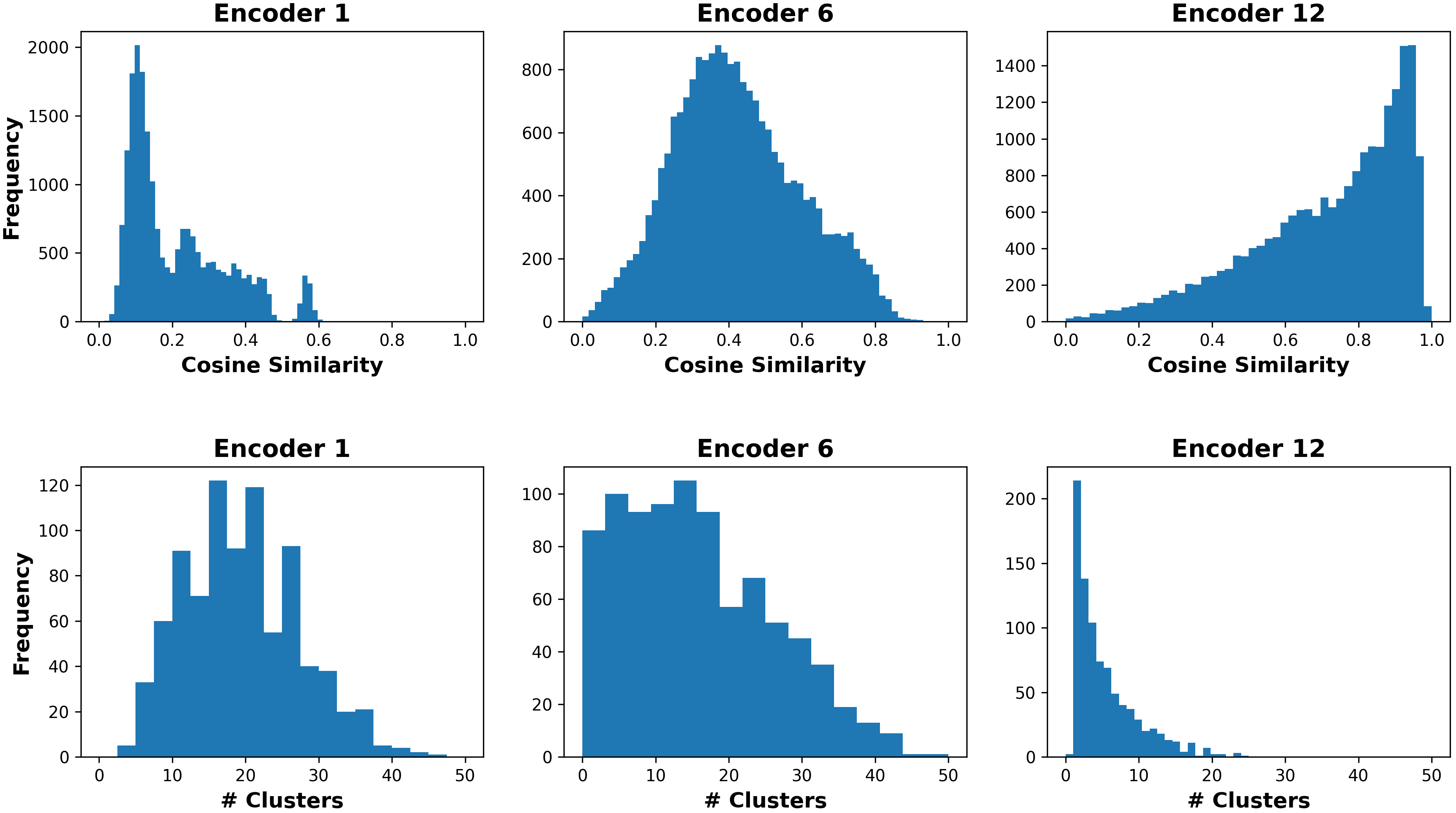}
\end{center}
\caption{SST-2 \textit{dev} set: Histogram of (a) similarity between CLS and all the other tokens (top row) (b) the number of clusters returned by DBSCAN (bottom row), over all the inputs at encoder $1$, $6$, and $12$.}
\label{fig:cls_similarity}
\end{figure}
In general, input sentences do not contain duplicate tokens. However, a preliminary study of token embeddings show that as the embeddings propagate through the pipeline of encoders, they become more similar to the CLS token, Figure  \ref{fig:cls_similarity} (top row). A deeper investigation shows that they also become more similar with each other and form clusters among themselves, Figure \ref{fig:cls_similarity} (bottom row). 


In this work, we exploit these observations, and present pyramid BERT, a novel BERT architecture, that reduces computational and space complexity of fine-tuning and inference of BERT while incurring minimal performance degradation. The pyramid BERT works for all the downstream tasks that only use the top layer CLS token for prediction, such as classification and ranking.

\begin{figure}
\begin{center}
  \includegraphics[width=.7\linewidth]{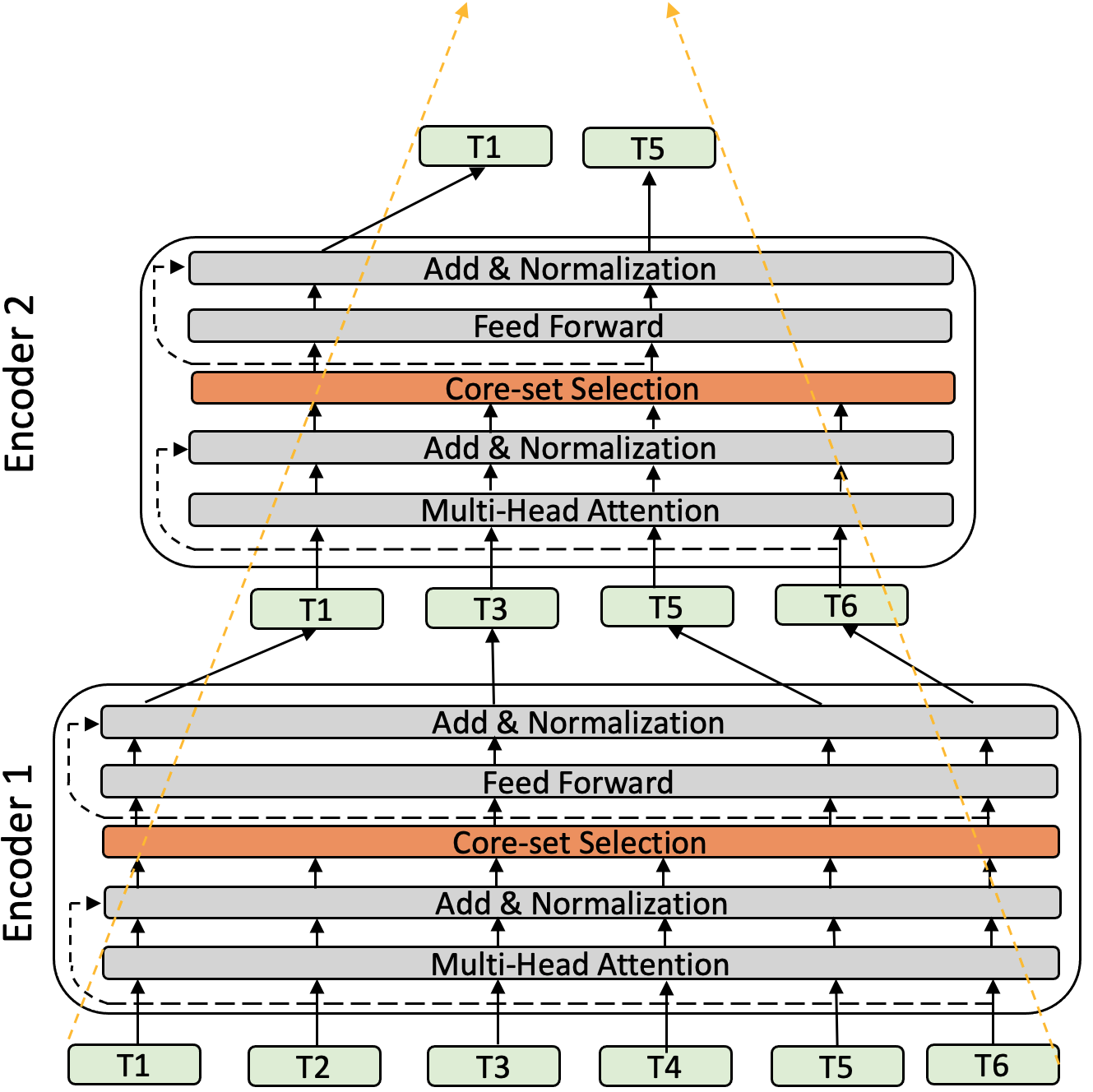}
\end{center}
  \caption{Illustration of Pyramid-BERT.}
  \label{fig:illustration-pyramid-bert}
\end{figure}

\textbf{Architecture.}
The pyramid BERT successively selects subsets of tokens in each encoder layer to propagate them to the next encoder. 
An illustration of pyramid BERT is shown in Figure \ref{fig:illustration-pyramid-bert}. It involves two main components over BERT: (a) A sequence-length configuration: a monotonically decreasing sequence ${\boldsymbol \ell} = (\ell_{1}, \ell_{2}, \cdots, \ell_{L})$ that specifies the number of tokens to retain in each of the $L$ encoders, for all the input examples.
(b) A core-set based token selection methodology which in each $j$-th encoder, given $\ell_{j-1}$ token embeddings from the $(j-1)$-th encoder selects a subset of it of size $\ell_j$ to propagate them to the next encoder.
Note that the rest of pyramid BERT architecture is same as the BERT, and it has exactly the same number of parameters as the BERT.

In Section \ref{sec:token_selection}, we provide a theoretical derivation of the core-set based token selection methodology by minimizing an upper bound of successive token selection loss. 
To computationally perform the core-set based token selection, we provide a greedy $k$-center algorithm in Section \ref{sec:token_algo}.
In Section \ref{sec:config}, we present a simple yet effective approach to select the sequence-length configuration $\boldsymbol \ell$ for a desired time/space complexity reduction.




\section{Coreset Based Token Selection}
\label{sec:token_selection}
\textbf{Problem Definition.}
We are interested in a $C$ class classification problem defined over a compact space $\mathcal{X}$ and a label space $\mathcal{Y} = \{1, 2, \cdots, C\}$. We consider a loss function $\mathcal{L}_{\bold{w}}(\cdot, \cdot) : \mathcal{X} \times \mathcal{Y} \rightarrow \mathcal{R}$ which is parametrized over the hypothesis class $(\bold{w})$, the parameters of the transformer network (e.g. BERT), and  
a set of training data points sampled i.i.d. over the space $\mathcal{Z} = \mathcal{X} \times \mathcal{Y}$ as $\{\bold{x}_i, y_i\}_{i \in [n]} \sim p_{\mathcal{Z}}$ where $[n] = \{1,2, \cdots, n\}$.



\begin{figure}
\begin{center}
  \includegraphics[scale=0.3]{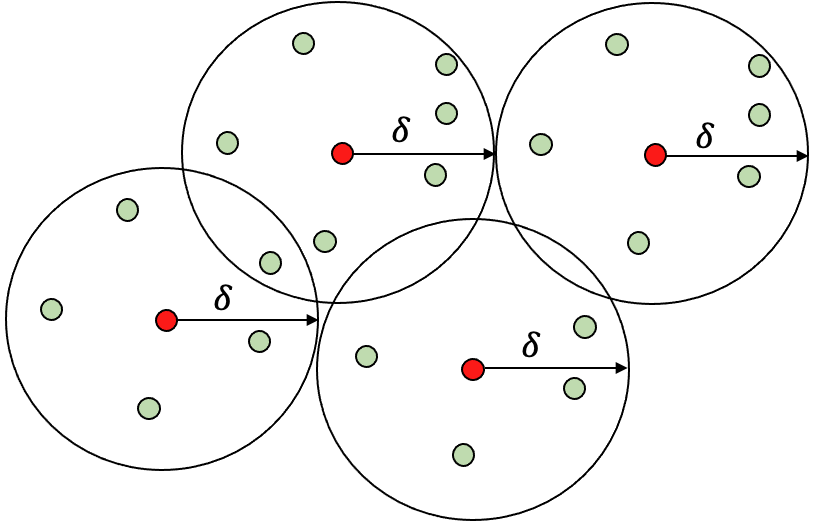}
\end{center}
  \caption{Illustration of $\delta$ cover core-set. 
  }
  \label{fig:illustration-core-set}
\end{figure}

Let $\mathcal{T}$ denote a token selection algorithm. For an example input $\bold{x}$, let the input and output embeddings of the token selection algorithm $\mathcal{T}$ at encoder $j$ be $\widetilde{S}_j$ and $S_j$ respectively.
The size of the two sets are $|\widetilde{S}_j| = \ell_{j-1}$, and $|S_j|= \ell_j$.
In particular, at each encoder $j$, the algorithm $\mathcal{T}$ selects $\ell_j$ embeddings of the input set $\widetilde{S}_j$ as the output set $S_j$, and eliminates the remaining  $\ell_{j-1}$ - $\ell_{j}$ embeddings in $\widetilde{S}_j$.
Let $\widetilde{\mathcal{S}} = \{\widetilde{S}_j\}_{j \in [L]}$, and ${\mathcal{S}} = \{{S}_j\}_{j \in [L]}$. Given the underlying BERT parameters $\bold{w}$, the sequence length configuration ${\boldsymbol \ell}$, and the classification loss $\mathcal{L}_{\bold{w}}$, pyramid BERT solves the token selection problem by minimizing the population risk as follows: 
\begin{align}\label{eq:prob}
\min_{\{\mathcal{S}: S_j \subseteq \widetilde{S}_j, |S_j| \leq \ell_j\}_{j\in [L]}} \mathbb{E}_{\bold{x},y\sim p_{\mathcal{Z}}}\big[\mathcal{L}_{\bold{w}}(\bold{x}, y, \widetilde{\mathcal{S}}, \mathcal{S})\big]\,.
\end{align}
\textbf{Method.} In order to design an optimal token selection algorithm $\mathcal{T}$, we consider the following upper bound of the token selection loss defined in \eqref{eq:prob}:
\begin{align}
\label{eq:upper_bound}
&\underbrace{\mathbb{E}_{\bold{x},y\sim p_{\mathcal{Z}}}[\mathcal{L}(\bold{x}, y, \widetilde{\mathcal{S}}, \mathcal{S})]}_{\text{pyramid BERT population risk}} \leq \underbrace{\frac{1}{n}\sum_{i=1}^{n} \mathcal{L}(\bold{x}_i, y_i)}_{\text{BERT training error}} + \nonumber \\
&\underbrace{\bigg| \mathbb{E}_{\bold{x},y\sim p_{\mathcal{Z}}}[\mathcal{L}(\bold{x}, y, \widetilde{\mathcal{S}}, \mathcal{S})] - \frac{1}{n}\sum_{i=1}^{n} \mathcal{L}(\bold{x}_i, y_i, \widetilde{\mathcal{S}}_i, \mathcal{S}_i) \bigg|}_{\text{pyramid BERT generalization error}} \nonumber \\
&+ \underbrace{\frac{1}{n}\sum_{i=1}^{n}\bigg| \mathcal{L}(\bold{x}_i, y_i) - \mathcal{L}(\bold{x}_i, y_i, \widetilde{\mathcal{S}}_i, \mathcal{S}_i)\bigg|}_{\text{pyramid BERT token selection loss}}\,.
\end{align}
For ease of notation, we write $\mathcal{L}_\bold{w}$ as $\mathcal{L}$.
The above bound follows immediately from the triangle inequality. 
%
For the first two terms in the above bound: the BERT training error is a constant for fixed parameters $\bold{w}$, and the generalization error of models like BERT is known to be small \citep{hao2019visualizing,jakubovitz2019generalization}. Therefore, we re-define the token selection problem, Equation \ref{eq:prob}, to minimize the third term, the pyramid BERT token selection loss
\begin{align}\label{eq:redfined_prob}
\frac{1}{n}\sum_{i=1}^{n}\min_{\{\mathcal{S}_i: |(S_{ij})| \leq \ell_j\}}  \bigg|\mathcal{L}(\bold{x}_i, y_i) - \mathcal{L}(\bold{x}_i, y_i, \widetilde{\mathcal{S}}_i, \mathcal{S}_i)\bigg|\,.
\end{align}
To solve Equation~\ref{eq:redfined_prob}, 
we first optimize a slightly different token selection algorithm $\mathcal{T}^*$ for a model pyramid$^*$ BERT. In pyramid$^*$ the embedding sequence length is not reduced across the encoder layers. 
Let $\widetilde{S}^*_j$ and $S^*_j$ denote the set of input and output embeddings of $\mathcal{T}^*$, where the size of the two sets is equal to the input sequence length $N$, $|\widetilde{S}^*_j| = |S^*_j|=N$. 
Given an input set $\widetilde{S}^*_j$ in encoder $j$, the algorithm $\mathcal{T}^*$ first selects a subset of it of size $\ell_j$, which is exactly same as $S_j$ selected by $\mathcal{T}$ in pyramid BERT. Next, instead of eliminating the remaining $N-\ell_j$ embeddings in $\widetilde{S}_j^*$ as is done by $\mathcal{T}$, the $\mathcal{T}^*$ replaces them with their nearest embedding in $S_j$ to form the output set $S^*_j$. The unique embeddings of the output set $S^*_j$ of pyramid$^*$ BERT are exactly same as the embeddings of output set $S_j$ of pyramid BERT, that is $\rm{unique}(S^*_j) = S_j$. We make the following observation. 
\begin{remark}
\label{rem:rem1}
A pyramid$^*$ BERT can be reduced to a pyramid BERT with a weighted self attention network. The weighted self attention network weighs attention scores according to the duplicity of the tokens in the embeddings of pyramid$^*$ BERT.
\end{remark}
Given the above remark, we optimize the token selection problem for pyramid$^*$ BERT. In the theorem below, we give an upper bound for the token selection loss stated in \eqref{eq:redfined_prob} for pyramid$^*$ BERT.
The proof relies on $\lambda$-Lipschitz continuity of the encoder and classification layers. 
A function $f$ is $\lambda$-Lipschitz continuity if, $\|f(x)-f(x')\| \leq \lambda \|x-x'\|$, for all $x, x' \in \rm{domain}(f)$.
\begin{theorem}
\label{thm:coreset_bound}
If the classification layer is $\lambda_C$-Lipschitz, and for all $j \in [L]$, the encoder $E_j$
is $\lambda_j$-Lipschitz continuous for all its parameters, and 
$\mathcal{T}^*$ is a token selection algorithm such that
the unique elements of the output embedding set 
$\rm{unique}(S_j^*)$ is a $\delta$-cover 
of the input embedding set
$\widetilde{S}_j^*$, and the $N- \ell_j$ remaining elements in $\widetilde{S}_j^* \setminus S_j^*$ are replaced in $S_j^*$
by their nearest elements in $\rm{unique}(S_j^*)$, 
then for all $i$ such that $\bold{x}_i$ is bounded, the following holds:
\begin{align}
&\Big|\mathcal{L}(\bold{x}_i, y_i) - \mathcal{L}(\bold{x}_i, y_i, \widetilde{\mathcal{S}}^*_i, \mathcal{S}^*_i)\Big| \nonumber\\
& \;\; \leq \delta \lambda_C \sum_{j=1}^L\bigg((N-\ell_j)\prod_{a=j}^{L}\lambda_{a}\bigg)\,.
\end{align}
\end{theorem}
We visualize the concept of $\delta$-cover in Figure \ref{fig:illustration-core-set}. 
The set of red points (i.e., token embeddings in our case) with radius $\delta$ covers the entire set of points. Theorem~\ref{thm:coreset_bound} suggests that we can bound the token selection loss of algorithm $\mathcal{T}^*$ for pyramid$^*$ BERT  with the $\delta$-cover core-set token selection. The loss goes to zero as the covering radius $\delta$ goes to zero. A proof of the theorem is given in Appendix \ref{app:proof}. 
%

From Remark \ref{rem:rem1}, the optimal choice of token selection for pyramid BERT is same as the optimal choice of unique tokens selected in pyramid$^*$ BERT, up to weighing of self-attention. 
However, in numerical experiments we found that fine-tuning pyramid BERT with the core-set based token selection performs better than weighing the self-attention. 
The $\delta$-cover core-set token selection problem is equivalent to the $k$-Center problem (also called min-max facility location problem) \cite{wolf2011facility}.
We explain how we solve the k-Center problem using a greedy approximation algorithm in \S\ref{sec:token_algo}.


\section{Pyramid BERT: Algorithm}
\label{sec:pyramid_algo}
Given a pre-trained BERT, we create a fine-tuned pyramid BERT as follows. For the core-set selection module shown in Figure \ref{fig:illustration-pyramid-bert}, we implement a $k$-Center-greedy-batch-$m$ algorithm, Section \ref{sec:token_algo}, to approximately select the core-set of embeddings. 
We fine-tune all the trainable parameters of the pyramid BERT for different choices of sequence-length configurations $\boldsymbol{\ell}$, according to approach given in Section \ref{sec:config}. We select the optimal configuration $\boldsymbol{\ell}$ satisfying the required inference speed-up or the space complexity reduction. The selected optimal choice of $\boldsymbol{\ell}$ is kept fixed during inference. 
We note that our practical implementation of pyramid BERT, proposed here, is not exactly the same as the theoretical token selection algorithm analyzed in the previous section. In Appendix \ref{app:algo_diff}, we justify the differences between the two.


\subsection{Token Selection Algorithm}
\label{sec:token_algo}
\begin{algorithm}[h]
{
\small
\caption{$k$-Center-greedy-batch-$m$}
\label{algo:algo1}
\KwData{Input set $\widetilde{S}$, the number of centers to add per iteration $m$.} 
\KwResult{Output set $S$, with $|S|=k$. }
Initialize $S = \{\textit{\rm CLS embedding}$\}\\
\While{$|S| < k$}{
    $M = \{\}$\\
    \While{$|M| \leq m$}{
    $s = \arg \max_{u \in \widetilde{S}\setminus S} \min_{v \in S}\textit{distance}(u,v)$\\
    $M = M \cup \{s\}$}
    $S = S \cup M$
}
\Return $S$
}
\end{algorithm}
The $\delta$-cover core-set problem is equivalent to $k$-Center problem which is NP-Hard. However, it is possible to obtain a $2 \times\rm{OPT}$ solution of $k$-Center using a greedy approach \cite{cook2009combinatorial}. 
The greedy approach selects the core-set of size $k$ one-by-one, making it un-parallelizable, and hence runs slow on GPUs.
For pyramid BERT, we developed a parallelizable version of this greedy approach, Algorithm \ref{algo:algo1},  which selects $m$ centers at a time. 




\subsection{Sequence-length Configuration}
\label{sec:config}
For the sequence-length configuration $\boldsymbol{\ell}$, we restrict the sequences to be exponentially decaying. Specifically, a valid sequence is determined by two parameters. The target pruning ratio $0 < p < 1$, and the index of the layer after which we stop reducing the sequence length $1 \leq i_\text{prune-upto} \leq L$. The sequence lengths are defined as 
\begin{equation}\label{equ: retention-config-decay}
    l_j = \bigg\lceil N \cdot p^{\frac{\min(j,i_\text{prune-upto}) }{i_\text{prune-upto}}}\bigg\rceil, j = 0, 1,\cdots, L,
\end{equation}
where $j=0$ corresponds to the input layer. We found that this strategy provides a good balance between the need to reduce the length quickly while retaining information.
It involves hyperparameter tuning with two HPs: $p, i_\text{prune-upto}$, and allows for an efficient training procedure.
In Appendix~\ref{app: sequence-len-generation-func} we provide a study comparing this restriction to possible alternatives showing its advantages. In addition we discuss how it compares to recent approaches~\citep{goyal2020power,ye2021tr}.

\section{Experiments}\label{sec: experiments}
We plug-in our \textit{core-set} based token selection method and the other competitive methods into encoder layers of a backbone Transformer, and after fine-tuning evaluate their performance on a wide range of natural language classification tasks. Specifically, we conduct the evaluations on two popular benchmarks: (1) the General Language Understanding Evaluation (GLUE) bench-mark~\citep{wang2018glue}, and (2) the Long Rang Arena (LRA), a collection of challenging long range context tasks~\citep{tay2020long}. For the backbone Transformer, we choose $\text{BERT}_\text{Base}$~\citep{devlin2018bert} for the GLUE benchmarks, and two state-of-the-art long-text Transformers \textit{Big Bird}~\citep{zaheer2020big} and \textit{Performers}~\citep{choromanski2020rethinking} for the LRA. For dataset statistics such as the number of classes and input sequence length, and the details of the backbone Transformers, see Appendix~\ref{app: experimental-setup}.

For sequence-length configurations, we generate a set $\mathcal{F}$ of $30$ sequence-length configurations using Equation~\ref{equ: retention-config-decay},
the details of which are listed in Table~\ref{tab: retention-config-specifics} Appendix~\ref{app: experimental-setup}. 
The high level idea is to cover multiple tradeoffs between efficiency and accuracy.

For each token selection method, we show the predictive performance for $1.5X$, $2X$, $3X$, and $3.5X$ speedup.
Similarly, we show the predictive performance for $30\%$ and $70\%$ space complexity reductions of the attention layer. 
The reason we consider the space reduction for the attention layer alone is that its quadratic complexity serves the main bottleneck for long sequences. 
For details on how to get the 
performance at different speedup and mathematical formula for computing speedup and space complexity reduction, see Appendix~\ref{app: experimental-setup}.

\subsection{Baseline Methods}
We compare our method with following five methods: (1) \textit{Attention-select} ({\bf Att}): An attention based mechanism from~\citet{goyal2020power}. (2) \textit{Average-pool} ({\bf Pool}): A strided mean pooling
applied to each sliding window of the sequence ~\cite{dai2020funnel}.
(3) \textit{First-k-select} ({\bf 1st}): Selects the first $k$ tokens, a strategy often considered with long documents.
(4) \textit{Input-first-k-select} ({\bf 1st-I}): Selects the first $k$ tokens, but rather than gradually reducing the sequence length in the encoder layers, performs a single truncation directly on the input. 
(5) \textit{Random-select} ({\bf Rand}): Selects a random subset of tokens. For all of the methods (including ours), the CLS token is always retained during the token selection.

For our \textit{core-set} based token selection, we try $6$ values of $m \in \{1, \lceil0.1 k\rceil, \lceil0.2 k\rceil, \lceil0.3 k\rceil, \lceil0.4 k\rceil, (k-1) \}$ where $k=l_j$, for $j=1,2,\cdots,L$. In particular, $m = k-1$ selects all $k-1$ tokens (centers) in one iteration given the first selected token is always the CLS token. And $m=1$ selects one token (center) per iteration which corresponds to the most fine-grained but also the slowest token selection strategy. We denote the strategy of $m=k-1$ as \textit{Coreset-select-k-1} ({\bf CS-$k$-1}), and the others as \textit{Coreset-select-x} where $\text{x} \in \{1, 0.1, 0.2, 0.3, 0.4\}$. We use \textit{Coreset-select-opt} ({\bf CS-opt}) to represent the best value from the \textit{Coreset-select-k-1} and \textit{Coreset-select-x}. In what follows, in tables presenting results we refer to the methods by their shortened (bold) names.

\subsection{Implementation}
To fairly evaluate our method against the baselines, we use the same set of hyperparameters for all the methods, for a given dataset. For details, see Appendix~\ref{app:hyperparameters}. 
The code for \text{Pyramid-BERT} is made available as a supplementary material with the submission. 
The training and inference jobs are run separately on a NVIDIA Tesla V$100$ GPU machine and a Intel Xeon Platinum $8000$ series CPU machine respectively. All the accuracy and speedup scores are averaged over $20$ trials.

\subsection{Results on GLUE benchmarks}\label{sec: results on glue datasets}
We first examine the trade-off between accuracy and speedup. The accuracy results for $3X$ and $1.5X$ speedup are summarized in Table~\ref{tab: glue-acc-speedup-3x} and~\ref{tab: glue-acc-speedup-1.5x} respectively. The results for $3.5X$ and $2X$ speedup are given in the Table \ref{tab: glue-acc-speedup-3.5x} and~\ref{tab: glue-acc-speedup-2x} in Appendix~\ref{app:glue-experiments-results}. We observe that as the speedup increases the gap between our \textit{Coreset-select-opt} and its competitors becomes large, where for $3X$ speedup, $\textit{Coreset-select-opt}$ outperforms the second best method \textit{Attention-select} by $1\%$ accuracy in average and beats the standard baselines by $2\%$ or more. The \textit{Average-pool} performs the worst in average across the GLUE benchmarks, specially on the COLA dataset. For detailed justification, see Appendix \ref{app:glue-experiments-results}.
To better understand the performance of $\textit{Coreset-select-opt}$ with different values of $m$, an ablation study is shown in Section~\ref{sec: ablation study}. 
For mild speedup of $1.5X$, we note that all methods (except \textit{Average-pool}) suffer only a small loss in accuracy and our method suffers no loss.
A similar situation occurs when viewing the tradeoff between space complexity and accuracy, where we provide results for a memory reduction of $70\%$ and $30\%$ in the Tables~\ref{tab: glue-acc-space-70} and~\ref{tab: glue-acc-space-30} (in \S~\ref{app:glue-experiments-results}).

\begin{table}[h]
\centering
\scalebox{0.56}{
 \begin{tabular}{l|ccccccc|c}
\toprule
 Dataset & 1st-I &   1st &  Rand & Pool &  Att & CS-$k$-1 & CS-opt &  $\text{BERT}_\text{Base}$ \\
\midrule
   STS-B &  $86.4$ &  $86.4$ &  $86.8$ & $81.6$ &  $\mathbf{87.0}$ &   $\mathbf{87.0}$ &   $\mathbf{87.0}$ &  $87.9$ \\
    MRPC &  $81.4$ &  $80.9$ &  $83.2$ & $83.9$ & $84.6$ &   $86.2$ &   $\mathbf{86.9}$ &  $87.3$ \\
   SST-2 &  $83.8$ &  $84.4$ &  $85.6$ & $85.2$ & $86.0$ &   $87.3$ &   $\mathbf{89.6}$ &  $92.4$ \\
    QNLI &  $84.8$ &  $84.4$ &  $86.4$ & $84.1$ & $86.8$ &   $\mathbf{87.8}$ &   $\mathbf{87.8}$ &  $90.9$ \\
    COLA &  $49.7$ &  $49.7$ &  $49.5$ & $3.0$ & $51.1$ &   $51.7$ &   $\mathbf{52.8}$ &  $53.3$ \\
     RTE &  $63.5$ &  $63.5$ &  $62.1$ & $59.2$ & $63.4$ &   $\mathbf{63.7}$ &   $\mathbf{63.7}$ &  $65.8$ \\
  MNLI\_M &  $77.8$ &  $76.7$ &  $81.4$ & $75.4$ & $\mathbf{82.5}$ &   $82.4$ &   $\mathbf{82.5}$ &  $84.0$ \\
 MNLI\_MM &  $75.9$ &  $75.6$ &  $78.7$ & $76.7$ & $\mathbf{82.7}$ &   $82.6$ &   $\mathbf{82.7}$ &  $84.6$ \\
     QQP &  $80.8$ &  $80.4$ &   $87.0$ & $79.4$ & $\mathbf{87.3}$ &    $\mathbf{87.3}$ &   $\mathbf{87.3}$ &  $87.5$ \\
\midrule
    Mean &  $76.0$ &  $76.1$ &  $77.9$ & $69.6$ & $79.0$ &   $79.6$ &    $\mathbf{80.0}$ &  $81.5$ \\
\bottomrule
\end{tabular}
}
\caption{\footnotesize GLUE \textit{dev} performance at $3X$ speedup. Here and everywhere else, F$1$ scores are reported for QQP and MRPC, Spearman correlations are reported for STS-B, Matthew's correlations are reported for
COLA, and
accuracy scores are reported for the other tasks. 
Each value is averaged over $20$ trials. Larger values indicates better performance. 
} 
\label{tab: glue-acc-speedup-3x}

\end{table}

\begin{table}
\centering
\scalebox{0.56}{
 \begin{tabular}{l|ccccccc|c}
\toprule
 Dataset & 1st-I &   1st &  Rand & Pool & Att & CS-$k$-1 & CS-opt &  $\text{BERT}_\text{Base}$ \\
\midrule
   STS-B &  $\mathbf{87.9}$ &  $\mathbf{87.9}$ &  $87.8$ & $87.8$ &  $\mathbf{87.9}$ &   $87.7$ &   $\mathbf{87.9}$ &  $87.9$ \\
    MRPC &  $86.8$ &  $86.4$ &  $87.2$ & $87.0$ & $87.1$ &   $86.9$ &   $\mathbf{87.3}$ &  $87.3$ \\
   SST-2 &  $92.1$ &  $91.5$ &  $91.9$ & $90.3$ & $92.3$ &   $\mathbf{92.4}$ &   $\mathbf{92.4}$ &  $92.4$ \\
    QNLI &  $90.8$ &  $90.8$ &  $90.8$ & $90.2$ & $90.7$ &   $\mathbf{90.9}$ &   $\mathbf{90.9}$ &  $90.9$ \\
    COLA &  $53.0$ &  $52.7$ &  $53.1$ & $25.6$ & $53.2$ &   $\mathbf{53.3}$ &   $\mathbf{53.3}$ &  $53.3$ \\
     RTE &  $65.6$ &  $65.2$ &  $\mathbf{65.7}$ & $61.5$ & $\mathbf{65.7}$ &   $65.4$ &   $65.6$ &  $65.8$ \\
  MNLI\_M &  $\mathbf{84.0}$ &  $83.8$ &  $83.9$ & $80.9$ & $\mathbf{84.0}$ &   $\mathbf{84.0}$ &   $\mathbf{84.0}$ & $84.0$ \\
 MNLI\_MM &  $84.1$ &  $84.0$ &  $83.9$ & $84.0$  & $84.5$ &   $\mathbf{84.6}$ &   $\mathbf{84.6}$ &  $84.6$ \\
     QQP &  $87.1$ &  $86.9$ &  $87.4$ & $85.7$ & $87.4$ &   $\mathbf{87.5}$ &   $\mathbf{87.5}$ &  $87.5$ \\
\midrule
    Mean &  $81.3$ &  $81.0$ &  $81.3$ & $76.8$ & $81.4$ &   $81.4$ &   $\mathbf{81.5}$ &  $81.5$ \\
\bottomrule
\end{tabular}
}
\caption{\footnotesize GLUE \textit{dev} performance at $1.5X$ speedup. 
}
\label{tab: glue-acc-speedup-1.5x}
\end{table}

\begin{table}
\centering
\scalebox{0.56}{
 \begin{tabular}{l|ccccccc|c}
\toprule
 Dataset & 1st-I &   1st &  Rand & Pool & Att & CS-$k$-1 & CS-opt &  $\text{BERT}_\text{Base}$ \\
\midrule
   STS-B &  $85.3$ &  $85.1$ &  $85.6$ & $78.7$ & $85.4$ &   $86.5$ &   $\mathbf{86.7}$ &  $87.9$ \\
    MRPC &  $81.3$ &  $81.5$ &  $83.3$ & $83.1$ & $84.3$ &   $86.0$ &   $\mathbf{86.6}$ &  $87.3$ \\
   SST-2 &  $83.3$ &  $84.6$ &  $84.9$ & $85.1$ & $87.2$ &   $87.6$ &   $\mathbf{87.7}$ &  $92.4$ \\
    QNLI &  $84.6$ &  $84.3$ &  $85.1$ & $84.0$ & $86.4$ &   $86.6$ &   $\mathbf{86.5}$ &  $90.9$ \\
    COLA &  $49.0$ &  $49.0$ &  $48.4$ & $0.0$ & $50.9$ &   $51.0$ &   $\mathbf{52.3}$ &  $53.3$ \\
     RTE &  $62.1$ &  $62.0$ &  $61.8$ & $59.8$ & $62.7$ &   $\mathbf{63.6}$ &   $\mathbf{63.6}$ &  $65.8$ \\
  MNLI\_M &  $76.9$ &  $76.3$ &  $79.0$ & $75.2$ & $80.5$ &   $80.9$ &   $\mathbf{81.0}$ & $84.0$ \\
 MNLI\_MM &  $74.9$ &  $74.5$ &  $79.3$ & $76.3$ & $80.7$ &   $81.6$ &   $\mathbf{81.8}$ &  $84.6$ \\
     QQP &  $80.6$ &  $80.0$ &  $86.6$ & $82.9$ & $87.0$ &   $87.2$ &   $\mathbf{87.3}$ &  $87.5$ \\
\midrule
    Mean &  $75.3$ &  $75.3$ &  $77.1$ & $69.5$  & $78.3$ &   $79.0$ &   $\mathbf{79.3}$ &  $81.5$ \\
\bottomrule
\end{tabular}
}
\caption{\footnotesize GLUE \textit{dev} performance at $70 \%$ space complexity reduction.
}
\label{tab: glue-acc-space-70}
\end{table}

\subsection{Results on Long Range Arena}
We show results on the following three datasets of LRA benchmark: (1) byte-level text classification using real-world data (IMDB), (2) Pathfinder task (long range spatial dependency problem), and (3) image classification on sequences of pixels converted from CIFAR-10. 


For baselines, we include \textit{First-k-select} and \textit{Random-select} methods, but fail to include \textit{Attention-select}. \textit{Attention-select} requires a full attention matrix for selecting tokens which is not available in \textit{Big Bird}~\citep{zaheer2020big} and \textit{Performers}~\citep{choromanski2020rethinking}. 
In addition, the Transformers including \textit{Big Bird} and \textit{Performers} in LRA have shallow architectures because of no pre-training: the default number of encoders for text classification, path finder, and image classification datasets are four, four, and one, respectively. Thus, for both baselines and our method, we only reduce sequence length in the \textit{input layer}, which is before the first encoder. For the sequence-length configurations, see Appendix~\ref{app:experiment-setup-lra}.
For \textit{Average-pool}, due to its worst performance on the GLUE benchmarks and the shallow architectures of the models in LRA, we exclude it from the baselines.

The results of accuracy scores for space complexity reduction at $70\%$ and $30\%$ are presented in Table~\ref{tab: long-text-70-reduction-all} and Table~
\ref{tab: long-text-30-reduction-all} (in Appendix~\ref{app:glue-experiments-results}), respectively. The \textit{Coreset-select-opt} here represents the \textit{Coreset-select} with $m=1$ because of its superior performance over other $m \in \{\lceil0.1 k\rceil\}, \lceil0.2 k\rceil, \lceil0.3 k\rceil, \lceil0.4 k\rceil\}$. 

We observe a similar pattern as discussed in GLUE benchmark evaluations: at high space complexity reduction $70\%$, \textit{Coreset-select-opt} significantly outperforms its competitors \textit{First-k-select} and \textit{Random-select} by $12\%$ and $2.5\%$ in average for \textit{Big Bird} ($12.3\%$ and $5.5\%$ in average for \textit{Performers}). Moreover, on CIFAR-10, our \textit{Coreset-select-opt} is even better than the \textit{Big Bird} and \textit{Performers} without any sequence reduction with accuracy gain $2.4\%$ and $2.6\%$, respectively (similarly for \textit{Performers} on PATHFINDER-32).
On the other hand, different from the GLUE evaluations, \textit{Coreset-select-k-1} does not show any significant advantages over the baseline methods. Our conjecture is that the input in the LRA datasets contain too many noisy or low level information which is not helpful for predicting the target. For an example, each pixel of an image (CIFAR-10) or a character in the byte-level text classification represents a token as the input. Our \textit{Coreset-select} based strategy with $m=1$ does the most fine-grained token-level selection than its baselines and thus filter out the noisy information. Note, we do not include accuracy and speedup tables because of insignificant gains observed in speedup due to the shallow architectures of Transformers in LRA.

\begin{table}
    \centering
    \scalebox{0.6}{
    \begin{tabular}{l|cccc|c}
    \toprule
    \multicolumn{6}{c}{\textit{Big Bird}} \\
    \midrule
    Dataset & 1st & Rand & CS-$k$-1 & CS-opt & Trans.-no-prune \\
    \midrule
    CIFAR-10 & $26.9$ & $39.4$ & $38.6$ & $\mathbf{43.3}$ & $40.9$ \\
    PATHFINDER-32 & $55.6$ & $69.9$ & $69.3$ & $\mathbf{71.7}$ & $73.5$ \\
    IMDB (BYTE-LEVEL) & $57.9$ & $59.6$ & $59.1$ & $\mathbf{61.4}$ & $63.8$\\
    \midrule
    Mean & $46.8$ & $56.3$ & $55.7$ & $\mathbf{58.8}$ & $59.4$ \\
    \midrule
        \multicolumn{6}{c}{\textit{Performers}} \\
    \midrule
        CIFAR-10 & $26.9$ & $41.5$ & $39.8$ & $\mathbf{45.5}$ & $42.9$ \\
    PATHFINDER-32 & $52.4$ & $58.2$ & $61.5$ & $\mathbf{67.7}$ & $66.2$ \\
    IMDB (BYTE-LEVEL) & $59.9$ & $59.9$ & $59.7$ & $\mathbf{62.8}$ & $64.3$\\
    \midrule
    Mean & $46.4$ & $53.2$ & $53.7$ & $\mathbf{58.7}$ & $57.8$ \\
    \bottomrule
    \end{tabular}
    }
    \caption{\footnotesize LRA \textit{test} set performances at $70 \%$ space complexity reduction for \textit{Big Bird} (top) and \textit{Performers} (bottom) as the backbone Transformer. 
    Here and everywhere else, accuracy scores are reported for all three tasks. Each value is averaged over $20$ trials. Larger values indicates better performance. 
    }
\label{tab: long-text-70-reduction-all}
    \label{tab:my_label}
\end{table}

\begin{table}
\centering
\scalebox{0.6}{
\begin{tabular}{ll|cc|cc|cc}
\toprule
\multicolumn{8}{c}{MNLI\_M} \\
\midrule
        \multicolumn{2}{l|}{Seq.len.config.} &  \multicolumn{2}{c|}{CS-1} &   \multicolumn{2}{c|}{CS-0.5} &  \multicolumn{2}{c}{CS-$k$-1} \\
\midrule
 $i_\text{prune-upto}$ & $p$ & Acc. & Speedup & Acc. & Speedup & Acc. & Speedup  \\ 

\midrule
          $1$ & $1.5$     &                       $\mathbf{78.3}$ &                            $2.5X$ &                              $77.9$ &                              $2.9X$ &                              $77.9$ &                              $3.6X$ \\
          $2$ & $0.2$ &                            $\mathbf{81.0}$ &                            $1.9X$ &                              $80.7$ &                              $2.3X$ &                              $80.6$ &                             $2.6X$ \\
         $3$ & $0.3$ &                            $\mathbf{82.9}$ &                            $1.7X$ &                              $82.8$ &                              $1.9X$ &                              $82.8$ &                              $2.1X$ \\
         $4$ & $0.4$ &                            $\mathbf{83.6}$ &                            $1.7X$ &                              $83.5$ &                              $1.7X$ &                              $83.4$ &                              $1.8X$ \\
\midrule          
\multicolumn{2}{c|}{$\text{BERT}_\text{Base}$} &                            $84.0$ &                            -- &                              $84.0$ &                              -- &                              $84.0$ &                              -- \\
\midrule
\multicolumn{8}{c}{MNLI\_MM} \\
\midrule
          $1$ & $1.5$&                           $\mathbf{79.2}$ &                            $2.4X$ &                              $78.3$ &                              $2.9X$ &                              $77.9$ &                              $2.9X$ \\
          $2$ &  $0.2$&                          $\mathbf{81.5}$ &                            $1.7X$ &                              $81.3$ &                              $2.2X$ &                              $81.2$ &                              $2.2X$ \\
          $3$ &   $0.3$&                         $\mathbf{83.5}$ &                            $1.2X$ &                              $83.4$ &                              $1.7X$ &                              $83.1$ &                              $1.7X$ \\
          $4$ &  $0.4$&                          $84.0$ &                            $1.1X$ &                              $\mathbf{84.1}$ &                              $1.4X$ &                              $84.0$ &                              $1.4X$ \\
\midrule          
\multicolumn{2}{c|}{$\text{BERT}_\text{Base}$} &                         $84.6$ &                       -- &                            $84.6$ &                         -- &                            $84.6$ &                         -- \\

\bottomrule
\end{tabular}
}
\caption{\textit{dev} set performance comparisons for \textit{Coreset-select} with $m \in \{1, \lceil0.5 k\rceil, k-1\}$. A fixed set of four sequence-length configurations are specified based on $i_\text{prune-upto}$ and $p$. 
}
\label{tab: k-centers-different-m}
\end{table}

\section{Ablation Studies}\label{sec: ablation study}
We conduct four ablation studies to better study pyramid-BERT: (1) Performance comparisons for \textit{Coreset-select} with different values of $m$, the number of centers to add per iteration. (2) Justification on the token importance measured by the \textit{Coreset-select}. 
(3) Comparison of applying \textit{Coreset-select} at both fine-tuning and inference versus at only inference to justify the necessity of fine-tuning in selecting tokens. (4) Exploring the position to plug-in the \textit{Coreset-select} in the encoder. 


The result for the first ablation study is presented in Table~\ref{tab: k-centers-different-m} 
We can see that \textit{Coreset-select} with $m=1$ gives the best performance but with the smallest speedup for MNLI-M/MM datasets.

Next, we conduct a study to validate the importance of tokens measured by our \textit{Coreset-select} strategy. We consider a trained BERT that has been fine-tuned on a downstream dataset without any sequence length reduction. During inference, given a encoder $j$ and input example consist of a sequence of tokens, we eliminate the $k$-th most important token measured by the \textit{core-set} selection method with $1\leq k \leq L$, and obtain a classification output (prediction label). The classification outputs for all input examples are then compared with those generated by BERT without any sequence length reduction. The comparison between the two classification outputs is measured by the mutual information~\citep{shannon2001mathematical}. Larger mutual information indicates more similarity between the two classification outputs. The importance score is specifically the order of tokens (centers) added by the \textit{core-set} selection method. For a batch of size $m$ tokens that are added at the same time, their importance is determined by the maximum distance between the token and its nearest centers that have already been added. The expectation is that the importance of tokens is negatively correlated with the corresponded mutual information. For an example, reducing the least important token for each input example should make the classification outputs have the least difference from those generated without reducing any token, and thus result in the largest mutual information.

\begin{figure}
\begin{center}
\includegraphics[scale=0.35]{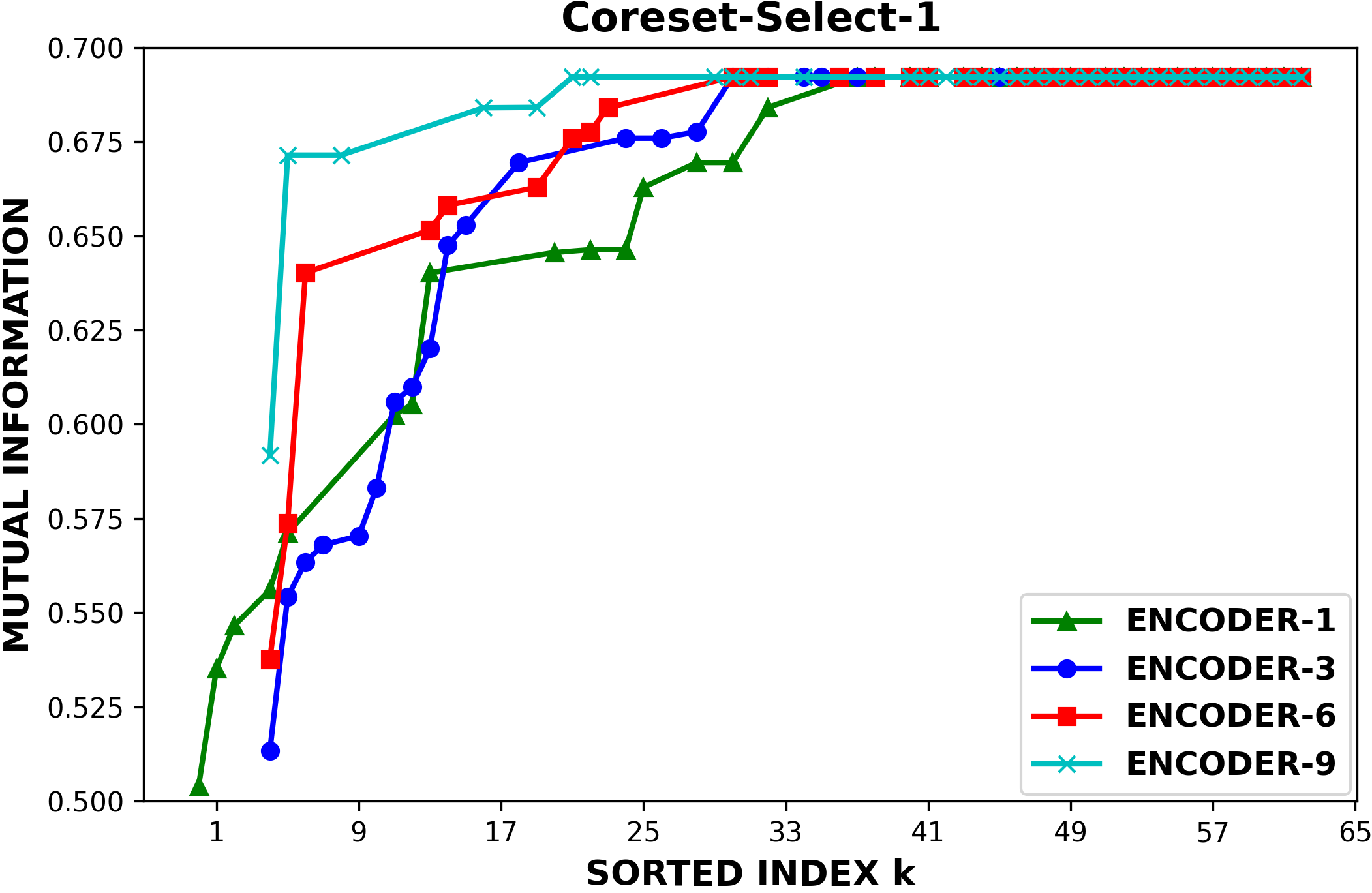}
\end{center}
\caption{Demonstration of token importance measured by the \textit{Coreset-select-1} based token selection method on SST-2 \textit{dev} set.}\label{fig: mutual info}
\end{figure}

\begin{figure}
\begin{center}
\includegraphics[scale=0.35]{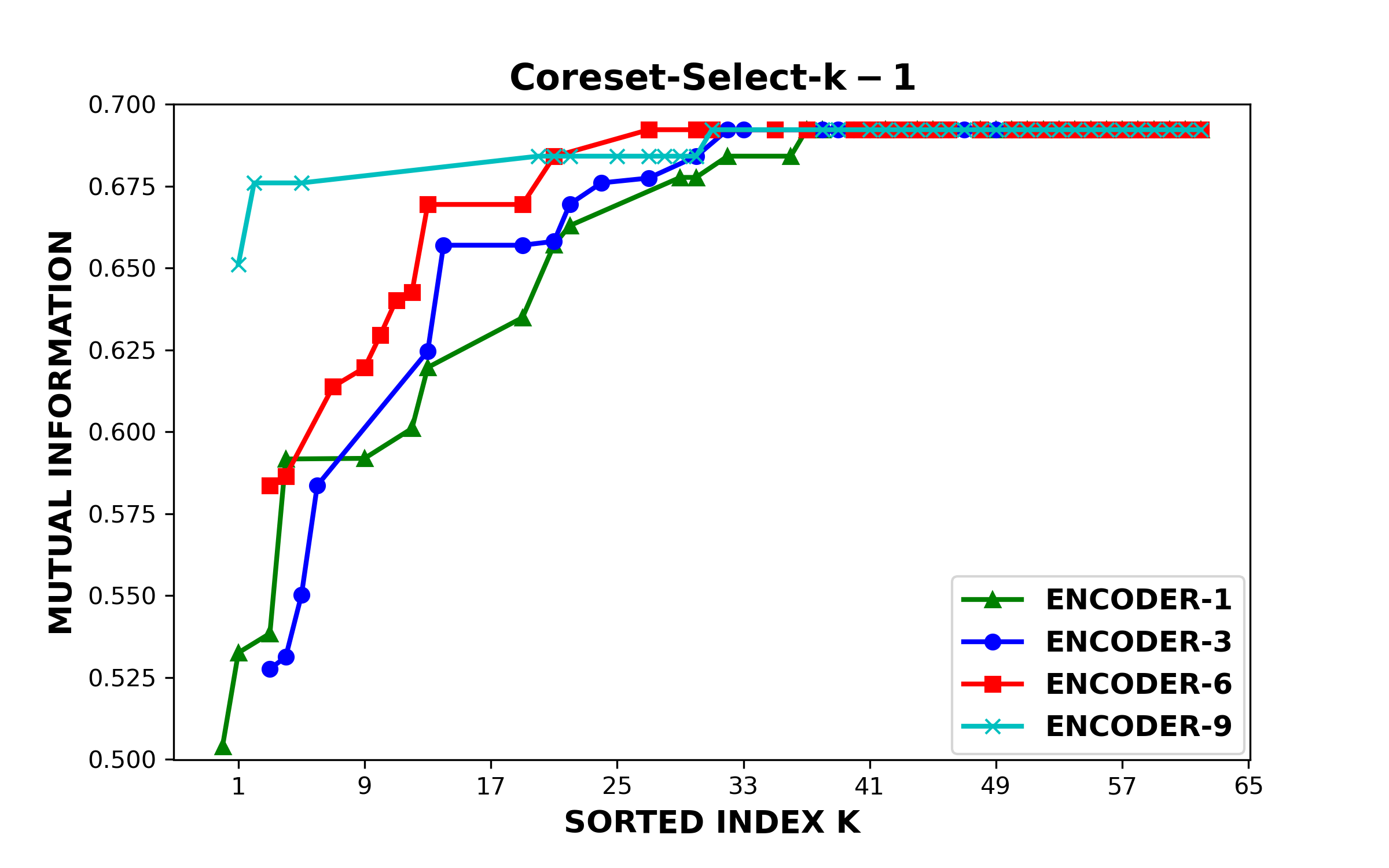}
\end{center}
\caption{Demonstration of token importance measured by the \textit{Coreset-select-k-1} based token selection method on SST-2 \textit{dev} set. }\label{fig: mutual info-cls}
\end{figure}

A result on SST-2 dataset is presented in Figure~\ref{fig: mutual info}. The input sequence length $L$ is set as $64$, and thus $k \in [1, 64]$. The encoder index $j$ takes value $\{1, 3, 6, 9\}$. Since the target variable of SST-2 is a relatively balanced binary value, the largest mutual information, which corresponds to no difference between the classification outputs, is $ln(2)\approx0.69$. For the token selection method, we choose \textit{Coreset-select} with $m=1$. The pattern shown in the figure aligns with our expectation that the importance of tokens is negatively correlated with the mutual information. Same pattern is observed for \textit{Coreset-select-k-1}. See Figure~\ref{fig: mutual info-cls}.


Next, we study the difference between applying \textit{Coreset-select} at both fine-tuning and inference, and at only inference. Table~\ref{tab:prune-train-no-train} justifies the necessary of fine-tuning in sequence length reduction.


\begin{table}
\centering
\scalebox{0.9}{
\begin{tabular}{l|cc}
\toprule
 & Only Infer & Fine-tune \& Infer \\
\midrule
QNLI & $68.2$ & $\mathbf{85.9}$ \\
SST-2 & $81.5$ & $\mathbf{87.2}$ \\
COLA & $47.9$ & $\mathbf{50.5}$ \\
MRPC & $83.5$ & $\mathbf{85.8}$ \\
STS-B & $84.6$ &  $\mathbf{86.5}$ \\
MNLI-M & $75.0$ & $\mathbf{80.6}$ \\
MNLI-MM & $74.8$ & $\mathbf{81.4}$\\
\bottomrule

\end{tabular}}
\caption{Comparison of token selection at inference only versus at both fine-tuning and inference. The particular sequence-length configuration is generated using Equation~\ref{equ: retention-config-decay} with $i_\text{prune-upto}$ as $3$, $p$ as $0.2$, and the input sequence length $N$ is $128$.}
\label{tab:prune-train-no-train}
\end{table}

Finally, we study the position to insert the \textit{Coreset-select} method in the encoder.
Two choices of position have been considered. The first option is to plug-in the token selection method right after the attention layer and before the feed-forward network, and the second option is to place it at the end of the encoder. 
The results is shown in Table~\ref{tab: where-to-insert}. The experiment shows that placing the token selection method right after the attention layer gives better performance than placing it at the end of encoder.


\begin{table}
\centering
\scalebox{0.9}{
\begin{tabular}{l|cc}
\toprule
 & Middle & End \\
 \midrule
MRPC & $\mathbf{85.8}$  & $84.0$  \\
RTE &  $\mathbf{63.4}$ & $60.4$ \\
MNLI-M & $\mathbf{80.7}$ & $79.9$\\
MNLI-MM & $\mathbf{81.4}$ & $80.1$ \\
\bottomrule
\end{tabular}}
\caption{Comparison between placing the \textit{Coreset-select} method in the middle (right after the attention layer) and at the end of the encoder layer. The particular sequence-length configuration is generated using Equation~\ref{equ: retention-config-decay} with $i_\text{prune-upto}$ as $3$, $p$ as $0.2$, and the input sequence length $N$ as $128$.}\label{tab: where-to-insert}
\end{table}


\section{Conclusion}
We provide pyramid-BERT, a theoretically justified technique for sequence length reduction, achieving speedup and memory reduction for both training and inference of transformers, while incurring significantly less accuracy drop than competitive methods.
However, this technique can be applied only for classification and ranking tasks which use single embedding from the top layer for prediction. Also, our token selection approach requires fine-tuning the network. An interesting future study would be to eliminate the need of fine-tuning.


\clearpage

\bibliographystyle{acl_natbib}
\bibliography{acl}

\clearpage
\appendix
\section{Appendix: Theory}
\subsection{Difference between the Theoretical and Practical Algorithm}
\label{app:algo_diff}
    There are two main aspects in which our practical implementation of pyramid BERT token selection differ from its theoretical one, for which we have provided guarantees in Theorem 1. Below we explain the differences and justify them.
    
    \begin{enumerate}
        \item In the theoretical implementation of pyramid BERT token selection, for which we have provided guarantees in Theorem 1, the original BERT parameters, $w$, are kept fixed, and the self-attention is weighted according to the duplicity of the tokens in the corresponding pyramid$^*$ BERT. Whereas in the practical implementation we remove the weighing of self-attention and offset it by fine-tuning the original BERT parameters $w$. We make this deviation as we found that fine-tuning BERT not only obviates the need of weighing self-attention but also reduces the performance degradation incurred due to token selection. We note that it is intractable to analyze this deviation in theory as BERT fine-tuning is a non-convex optimization.
        \item In the theoretical implementation of pyramid BERT, a $\delta$-cover core-set of tokens is selected. The $\delta$-cover core-set token selection is equivalent to the $k$-Center problem. The theoretical guarantees assume that we can get the optimal solution of the $k$-Center problem. However, the $k$-Center problem is NP-hard and a best known algorithm of it $k$-Center-greedy gives a $2 \times \rm{OPT}$ solution. In our practical implementation, we go a step beyond the greedy approach and propose a parallelizable version of the $k$-Center-greedy algorithm that takes in an additional hyper-parameter $m$. The hyper-parameter $m$ sets the level of parallelization and the choice of $m=1$ reduces it to the original $k$-Center-greedy. In the numerical experiments, we report the accuracy for the best pyramid BERT by optimizing over the hyper-parameter $m$.
    \end{enumerate}
    We note that despite the above mentioned differences of our practical algorithm from the theoretical one, the theoretical guarantees achieved in Theorem 1 do justify the approach of core-set based token selection. The theorem establishes that the token selection loss of the pyramid BERT is bounded by the $\delta$-cover of the core-set. Informally, if the input sentence comprises near-duplicate tokens, in pyramid BERT, the loss incurred by the token selection method goes to zero.
\subsection{Proof of Theorem \ref{thm:coreset_bound}}
\label{app:proof}
On a high level, theorem follows from (1) the definition of Lipschitz continuity, (2) the definition of the token selection algorithm $\mathcal{T}^*$ of pyramid$^*$ BERT, and (3) the fact that the loss function $\mathcal{L}_{\bold{w}}$ comprises a sequence of encoder layers stacked on top of each other. 

We introduce two new notations. Let $\mathcal{O}_j$ denote the BERT model up to the output of the $j$-th encoder, and $\mathcal{E}_j$ denote the BERT network up to the input of the $j$-th encoder. We assume that the token selection algorithm $\mathcal{T}^*$ operates on the embeddings before they are inputted to the encoder. Also, for ease of notation we omit the subscript $i$ denoting the $i$-th training example from $(\bold{x}_i,y_i)$.

Based on the above notations, for BERT we have $\mathcal{E}_j(\bold{x},y) = \mathcal{O}_{j-1}(\bold{x},y)$. For pyramid$^*$ BERT, due to the token selection algorithm $\mathcal{T}^*$, for each $j$-th encoder layer, we have,
\begin{align}
\label{eq:ap0}
\bigg\|\mathcal{E}_{j}(\bold{x}, y, \widetilde{\mathcal{S}}^*, \mathcal{S}^*) - \mathcal{O}_{j-1}(\bold{x}, y, \widetilde{\mathcal{S}}^*, \mathcal{S}^*)\bigg\| \nonumber \\
\leq \delta (N- \ell_j)\,.
\end{align}
The above equation uses the fact that at most $(N- \ell_j)$ tokens are replaced with their corresponding core-set center token, which is at most $\delta$ away from them. The Equation \eqref{eq:ap1} follows immediately from the definition of Lipschitz continuity of the classification layer. 
\begin{align}
\label{eq:ap1}
&\Big|\mathcal{L}(\bold{x}, y) - \mathcal{L}(\bold{x}, y, \widetilde{\mathcal{S}}^*, \mathcal{S}^*)\Big| \nonumber\\ 
& \leq \lambda_{C} \Big\|\mathcal{O}_{L}(\bold{x}, y) - \mathcal{O}_{L}(\bold{x}, y, \widetilde{\mathcal{S}}^*, \mathcal{S}^*)\Big\|\,.
\end{align}

The Equation \eqref{eq:ap2} follows from the Lipschitz continuity of the $L$-th encoder. 
The Equation \eqref{eq:ap3}  follows from Equation \eqref{eq:ap0}. The Equation \eqref{eq:ap4} follows from the repeated application of Equation \eqref{eq:ap3} over the next encoder layer. The Equation \eqref{eq:ap5} follows from the repeated application of Equation \eqref{eq:ap3} over the subsequent encoder layers, and the fact that $\mathcal{O}_{0}(\bold{x}, y) = \mathcal{O}_{0}(\bold{x}, y, \widetilde{\mathcal{S}}^*, \mathcal{S}^*)$, as input to the first encoder layer is same for BERT and pyramid$^*$ BERT.

\begin{align}
&\Big\|\mathcal{O}_{L}(\bold{x}, y) - \mathcal{O}_{L}(\bold{x}, y, \widetilde{\mathcal{S}}^*, \mathcal{S}^*)\Big\|\nonumber\\
& \leq \lambda_{L} \Big\|\mathcal{E}_{L}(\bold{x}, y) - \mathcal{E}_{L}(\bold{x}, y, \widetilde{\mathcal{S}}^*, \mathcal{S}^*)\Big\|\label{eq:ap2} \\
& \leq \lambda_{L}\delta(N-\ell_L) \nonumber\\
&+  \lambda_{L}\Big\|\mathcal{O}_{L-1}(\bold{x}, y) - \mathcal{O}_{L-1}(\bold{x}, y, \widetilde{\mathcal{S}}^*, \mathcal{S}^*)\Big\| \label{eq:ap3} \\
& \leq \lambda_{L}\delta(N-\ell_L) + \lambda_L\lambda_{L-1}\delta(N-\ell_{L-1}) \nonumber\\
&+  \lambda_{L}\lambda_{L-1}\Big\|\mathcal{O}_{L-2}(\bold{x}, y) - \mathcal{O}_{L-2}(\bold{x}, y, \widetilde{\mathcal{S}}^*, \mathcal{S}^*)\Big\| \label{eq:ap4}\\
& \leq \lambda_{L}\delta(N-\ell_L) + \lambda_L\lambda_{L-1}\delta(N-\ell_{L-1}) \nonumber\\
&+  \cdots + \Big(\prod_{j=0}^{L-1}\lambda_{L-j}\Big)\delta (N-\ell_1) \label{eq:ap5}\\
& = \delta \sum_{j=1}^L\bigg((N-\ell_j)\prod_{i=j}^{L}\lambda_{i}\bigg)\,. \label{eq:ap6}
\end{align}
The theorem follows by combining Equations \eqref{eq:ap1} and \eqref{eq:ap6}.

\section{Appendix: Evaluation on the Sequence-length Generation Function}\label{app: sequence-len-generation-func}
We note that most recent approaches such as~\citep{goyal2020power,ye2021tr} try to learn a task-dependent sequence-length configuration with a cost of fine-tuning more than twice on the downstream data, where the first fine-tuning trains a full model without any sequence-length reduction, with additional parameters that often requires delicate tuning. This approach does not help the training process and in fact increase its cost whereas our goal is to improve the training procedure.
Furthermore, there is still an amount of HP tuning involved in order to find the right ratio of acceleration (or memory reduction) to accuracy.
Our technique involves hyperparameter tuning but with two HPs: $p, i_\text{prune-upto}$, and allows for an efficient training procedure.

\begin{figure}[h]
\begin{center}
\includegraphics[scale=0.35]{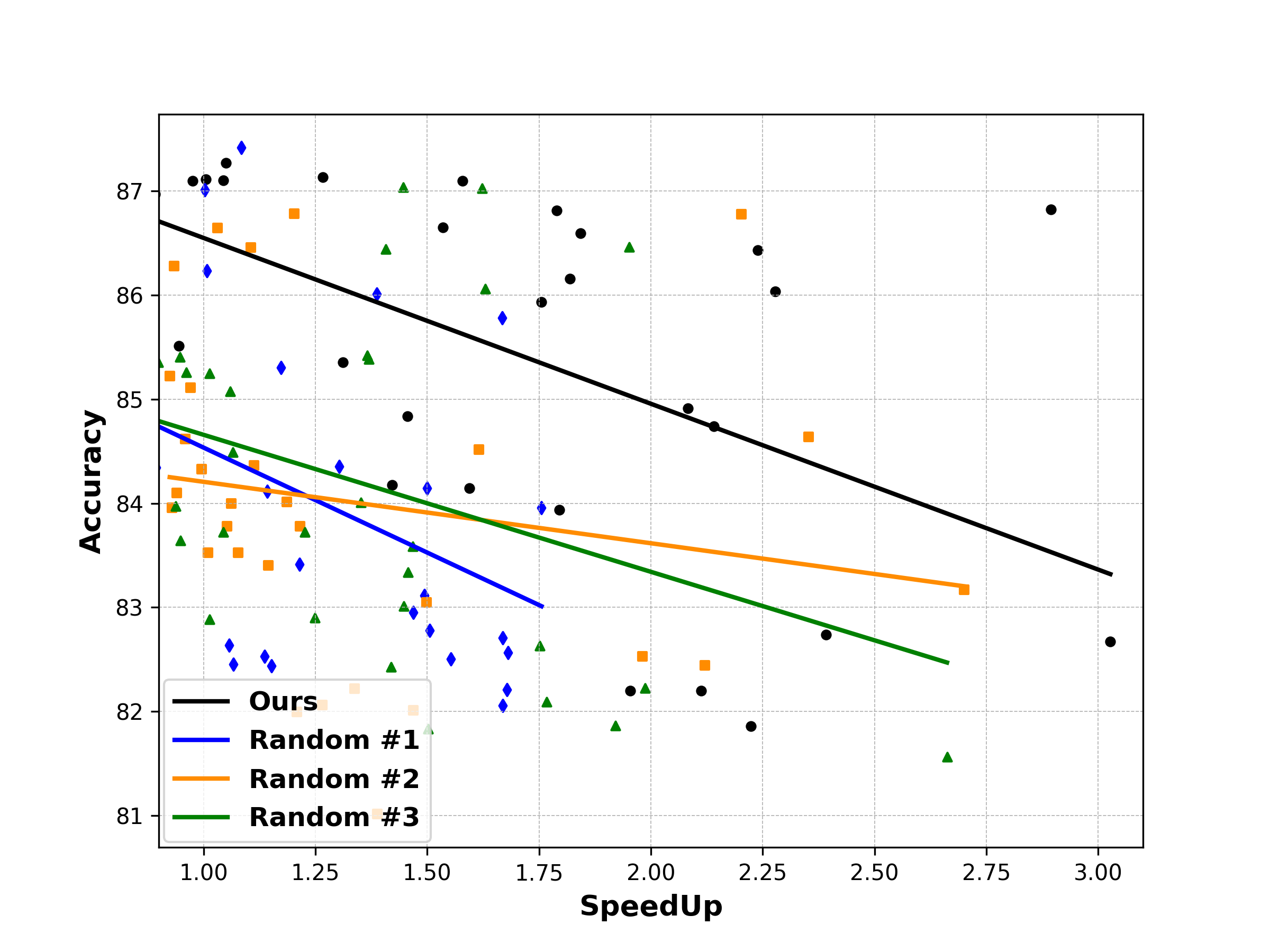}
\end{center}
\caption{MRPC: Scatter plots of accuracy versus speedup for $30$ sequence-length configurations generated by our and random method. Each solid line is from a linear regression model fitted on the scatter points of the corresponded method.}
\label{fig: box-plot}
\end{figure}

We conduct an experiment to validate the sequence-length generation function, in comparison to a random method that gives configurations for all encoders. Given a dataset we use the retention generation function  in Equation~\ref{equ: retention-config-decay} and random method to generate a fix number of sequence-length configurations, separately. Then we apply the same \textit{core-set} based method on the dataset with each sequence-length configuration and compute the statistics of accuracy and speedup for our and random method. We repeat the random method for three times. The number of sequence-length configurations is set as $30$, and details of those generated by the Equation~\ref{equ: retention-config-decay} is presented in Table~\ref{tab: retention-config-specifics} in Appendix~\ref{app: experiment-setup-glue}. The results for SST-2 dataset are shown in Figure~\ref{fig: box-plot}. We can see that the sequence-length configurations generated by our method provide wider searching range of accuracy and speedups than those generated by the random method. 

\section{Appendix: Experiments Setup}\label{app: experimental-setup}
Data statistics such as the number of classes and input sequence length are specified in Table~\ref{tab: data-statistics}.

\begin{table}
\centering
\scalebox{0.5}{
\begin{tabular}{lcc}
\toprule
               DATASET &  \# CLASSES & INPUT SEQUENCE LENGTH ($N$) \\
\midrule
                 STS-B &        --  &                       128 \\
                  MRPC &        2 &                      128 \\
                 SST-2 &        2 &                        64 \\
                  QNLI &        2 &                    128 \\
                  COLA &        2 &                     64 \\
                   RTE &        2 &                    256 \\
                MNLI-M &        3 &                    128 \\
               MNLI-MM &        3 &                    128 \\
                   QQP &        2 &                    128 \\
\midrule                   
               CIFAR10 &       10 &                  1024 \\
          PATHFINDER32 &        2 &                  1024 \\
 IMDB (BYTE-LEVEL) &        2 &                      1000 \\
\bottomrule
\end{tabular}}
\caption{Dataset statistics for GLUE and LRA benchmarks. STS-B is a regression task and thus does not have classes.}\label{tab: data-statistics}
\end{table}

The details of the backbone Transformer used in GLUE\citep{wang2018glue} and LRA~\citep{tay2020long} are presented as follows: For $\text{BERT}_\text{Base}$, it was pre-trained on the BooksCorpus and English Wikipedia with $L=12$ encoders, $A = 12$ self-attention heads per encoder and hidden size $H = 768$. For \textit{Big Bird}~\citep{zaheer2020big} and \textit{Performers}~\citep{choromanski2020rethinking}, we follow the original implementation in LRA~\citep{tay2020long} that models are fine-tuned from scratched on each task without any pre-training. 

\subsection{GLUE Benchmarks}\label{app: experiment-setup-glue}

For experiments on GLUE benchmarks, the sequence-length configurations $\mathcal{F}$ generated by Equation~\ref{equ: retention-config-decay} are presented in Table~\ref{tab: retention-config-specifics}. For each dataset and Transformer model with a token selection method, the same set of sequence-length configurations are used.
\begin{table*}
    \centering
    \scalebox{0.57}{
\begin{tabular}{cc|cccccccccccc}
    \toprule
 \multicolumn{2}{c|}{Input} &  \multicolumn{12}{c}{Output: Retention configurations -- Number of tokens to retain at each encoder layer}\\
 \midrule
\vtop{\hbox{\strut The last layer}\hbox{\strut index to apply}\hbox{\strut token selection}\hbox{\strut $i_\text{prune-upto}$}}  & \vtop{\hbox{\strut The proportion of}\hbox{\strut input tokens $p$ }\hbox{\strut to retain at $i_\text{prune-upto}$}} &  layer 1 &  layer 2 &  layer 3 &  layer 4 &  layer 5 &  layer 6 &  layer 7 &  layer 8 &  layer 9 &  layer 10 &  layer 11 &  layer 12 \\
\midrule
  2 &  0.15 &       49 &       19 &       19 &       19 &       19 &       19 &       19 &       19 &       19 &        19 &        19 &        19 \\
  3 &  0.15 &       68 &       36 &       19 &       19 &       19 &       19 &       19 &       19 &       19 &        19 &        19 &        19 \\
  4 &  0.15 &       79 &       49 &       30 &       19 &       19 &       19 &       19 &       19 &       19 &        19 &        19 &        19 \\
  3 &  0.10 &       59 &       27 &       12 &       12 &       12 &       12 &       12 &       12 &       12 &        12 &        12 &        12 \\
  4 &  0.10 &       71 &       40 &       22 &       12 &       12 &       12 &       12 &       12 &       12 &        12 &        12 &        12 \\
  3 &  0.20 &       74 &       43 &       25 &       25 &       25 &       25 &       25 &       25 &       25 &        25 &        25 &        25 \\
  4 &  0.20 &       85 &       57 &       38 &       25 &       25 &       25 &       25 &       25 &       25 &        25 &        25 &        25 \\
  3 &  0.17 &       70 &       39 &       21 &       21 &       21 &       21 &       21 &       21 &       21 &        21 &        21 &        21 \\
  3 &  0.18 &       72 &       40 &       23 &       23 &       23 &       23 &       23 &       23 &       23 &        23 &        23 &        23 \\
  3 &  0.19 &       73 &       42 &       24 &       24 &       24 &       24 &       24 &       24 &       24 &        24 &        24 &        24 \\
  3 &  0.22 &       77 &       46 &       28 &       28 &       28 &       28 &       28 &       28 &       28 &        28 &        28 &        28 \\
  3 &  0.25 &       80 &       50 &       32 &       32 &       32 &       32 &       32 &       32 &       32 &        32 &        32 &        32 \\
  1 &  0.25 &       32 &       32 &       32 &       32 &       32 &       32 &       32 &       32 &       32 &        32 &        32 &        32 \\
  2 &  0.25 &       64 &       32 &       32 &       32 &       32 &       32 &       32 &       32 &       32 &        32 &        32 &        32 \\
  3 &  0.25 &       80 &       50 &       32 &       32 &       32 &       32 &       32 &       32 &       32 &        32 &        32 &        32 \\
  5 &  0.25 &       97 &       73 &       55 &       42 &       32 &       32 &       32 &       32 &       32 &        32 &        32 &        32 \\
  9 &  0.25 &      109 &       94 &       80 &       69 &       59 &       50 &       43 &       37 &       32 &        32 &        32 &        32 \\
 11 &  0.25 &      112 &       99 &       87 &       77 &       68 &       60 &       52 &       46 &       41 &        36 &        32 &        32 \\
  1 &  0.50 &       64 &       64 &       64 &       64 &       64 &       64 &       64 &       64 &       64 &        64 &        64 &        64 \\
  2 &  0.50 &       90 &       64 &       64 &       64 &       64 &       64 &       64 &       64 &       64 &        64 &        64 &        64 \\
  3 &  0.50 &      101 &       80 &       64 &       64 &       64 &       64 &       64 &       64 &       64 &        64 &        64 &        64 \\
  5 &  0.50 &      111 &       97 &       84 &       73 &       64 &       64 &       64 &       64 &       64 &        64 &        64 &        64 \\
  9 &  0.50 &      118 &      109 &      101 &       94 &       87 &       80 &       74 &       69 &       64 &        64 &        64 &        64 \\
 11 &  0.50 &      120 &      112 &      105 &       99 &       93 &       87 &       82 &       77 &       72 &        68 &        64 &        64 \\
  1 &  0.75 &       96 &       96 &       96 &       96 &       96 &       96 &       96 &       96 &       96 &        96 &        96 &        96 \\
  2 &  0.75 &      110 &       96 &       96 &       96 &       96 &       96 &       96 &       96 &       96 &        96 &        96 &        96 \\
  3 &  0.75 &      116 &      105 &       96 &       96 &       96 &       96 &       96 &       96 &       96 &        96 &        96 &        96 \\
  7 &  0.75 &      122 &      117 &      113 &      108 &      104 &      100 &       96 &       96 &       96 &        96 &        96 &        96 \\
  9 &  0.75 &      123 &      120 &      116 &      112 &      109 &      105 &      102 &       99 &       96 &        96 &        96 &        96 \\
 11 &  0.75 &      124 &      121 &      118 &      115 &      112 &      109 &      106 &      103 &      101 &        98 &        96 &        96 \\
\bottomrule
\end{tabular}}
\caption{Sequence-length configurations $\mathcal{F}$ for experiments on GLUE benchmarks. Each sequence-length configuration is determinied by the configuration generation function (Equation~\ref{equ: retention-config-decay}) which requires two hyperparameters ``the last layer index to apply token selection $i_\text{prune-upto}$" and ``the proportion of input tokens $p$ to retain at $l_\text{prune-upto}$". The input sequence $N$ is set as $128$.}\label{tab: retention-config-specifics}
\end{table*}

For each sequence-length configuration in $\mathcal{F}$ a Transformer with a token selection method is applied at both fine-tuning and inference. 
Next, for each token selection method we select the accuracy scores when there are speedup $1.5X$, $2X$, $3X$, and $3.5X$ over the Transformers without any sequence reduction. To get the accuracy score at the exact speedup $X$, a linear interpolation is used when necessary. However, for objective evaluation we do not extrapolate the results. For details on how to get accuracy scores at different speedup number, see Figure~\ref{fig:demo-scatter-pareto-sst-2} and~\ref{fig:demo-scatter-pareto-mrpc} in Appendix~\ref{app:glue-experiments-results}. Similarly, we select the accuracy scores when the space complexity reductions for the attention layer are $30\%$ and $70\%$. The reason why we only focus the space reduction for the attention layer is that its quadratic complexity serves the main efficiency bottleneck for Transformers. 

For \textit{Input-first-k-select}, since it does not rely on $\mathcal{F}$ but different truncated input sequence lengths, we specify its configuration as following: For dataest with $N=256$, we try truncated sequence lengths of $\{240, 224, \cdots, 48, 32, 16\}$. For dataset with $N=120$, we try truncated sequence lengths of  $\{112, 96, 80, 64, 48, 32, 16, 8\}$. And for dataset with $N=64$, we try truncated sequence lengths of $\{48, 32, 16, 8, 4\}$.

Similarly, \textit{Average-pool} does not rely on $\mathcal{F}$ but the window size and encoder layer(s) to apply the pooling. To get the accuracy scores at various speedups, we try the average pooling with different window sizes of $\{2, 3, 4, 5, 6\}$ on various encoder layer(s). The window size is set equal to the stride size.

The mathematical formulas for speedup and space complexity reduction are presented as below. Consider $\text{BERT}_\text{Base}$ as the backbone Transformer for an example, and denote the $\text{BERT}_\text{Base}$ with a token selection method as $\text{BERT}_\text{Token-select}$. The speedup is computed as $ T(\text{BERT}_\text{Base}) \mathbin{/} T(\text{BERT}_\text{Token-select})$, where $T(\cdot)$ is the time duration in inference. Larger value indicates higher inference speedup for $\text{BERT}_\text{Token-select}$ over $\text{BERT}_\text{Base}$. The space complexity reduction for the attention layer is computed as $ 1 - S(\text{BERT}_\text{Token-select}) \mathbin{/} S(\text{BERT}_\text{Base})$, where $S(\cdot) =  \sum_{j=1}^{L} l_j^2 + l_j d, j = 1, 2, \cdots, L$, and $l_j$ and $d$ denotes the number of tokens to select at encoder $j$ and the hidden dimension (dimension of the latent representation), respectively. Larger value indicates higher space complexity reduction for $\text{BERT}_\text{Token-select}$ over $\text{BERT}_\text{Base}$. For simplicity, the ``space complexity reduction" refers to the reduction for the attention layer in the following discussion. 



\subsection{LRA}\label{app:experiment-setup-lra}
We run a set of sequence-length configurations where the number of tokens to select on the input layer is $\{\lceil0.1\cdot N\rceil, \lceil0.2\cdot N\rceil, \cdots, \lceil0.9\cdot N\rceil\}$ and $N$ is the input sequence length. The \textit{Coreset-select-opt} here represents the \textit{core-set} token selection method with $m=1$ because we observe it gives the best performance than $m \in \{\lceil0.1\cdot k\rceil\}, \lceil0.2\cdot k\rceil, \lceil0.3\cdot k\rceil, \lceil0.4\cdot k\rceil\}$. The results of accuracy scores for space complexity reduction at $70\%$ and $30\%$ are presented in Table~\ref{tab: long-text-70-reduction-all} and~\ref{tab: long-text-30-reduction-all}, respectively. We observe a similar pattern as shown in Section~\ref{sec: results on glue datasets}: at high space complexity reduction $70\%$ in Table~\ref{tab: long-text-70-reduction-all}, \textit{Coreset-select-opt} significantly outperforms its competitors \textit{First-k-select} and \textit{Random-select} by $12\%$ and $2.5\%$ in average for \textit{Big Bird} ($12.3\%$ and $5.5\%$ in average for \textit{Performers}). Moreover, on CIFAR-10, our \textit{Coreset-select-opt} is even better than the \textit{Big Bird} and \textit{Performers} without any token selection with accuracy gain $2.4\%$ and $2.6\%$, respectively. Similarly on PATHFINDER-32, the \textit{Coreset-select-opt} shows $1.5\%$ accuracy gain over the \textit{Performers} without any token selection. On the other hand, different from Section~\ref{sec: results on glue datasets}, \textit{Coreset-select-k-1} does not show any significant advantages over the baseline methods. Our conjecture is that the input in the LRA tasks contain too many noisy or low level information which is not helpful for predicting the target. For an example, each pixel of an image (CIFAR-10) or a character in the byte-level text classification represents a token in the input for the Transformer. Our \textit{core-set} based strategy, especially with $m=1$, does the most fine-grained token-level selection than its baselines and thus filter out the noisy information. 
At low space complexity reduction $30\%$ in Table~\ref{tab: long-text-30-reduction-all}, the advantages for \textit{Coreset-select-opt} over its baselines becomes smaller, which matches our expectations in the mild sequence-length reduction regime. Note, we do not include accuracy and speedup tables because of insignificant gains observed in speedup due to the shallow architectures of Transformers in LRA and the usage of the slowest  \textit{coreset} based method with $m=1$.

\section{Appendix: Hyperparameters}\label{app:hyperparameters}
To fairly evaluate our method, we do not tune any hyperparameters for both GLUE and LRA benchmarks, i.e., the same set of hyperparameters are used for each dataset across every competing method. For GLUE benchmarks, the learning rate and number of epochs are set as $2e^{-5}$ and $3$, respectively. The batch size in training is set as $48$ for all except MNLI-M/mm and RTE, which are set as $32$ and $16$, respectively. The batch size in inference is uniformly set as $128$. For LRA~\citep{tay2020long}, we follow the exact settings for the task-specific hyperparameters and models configurations provided in its official github repository, except reducing the number of encoders in the backbone Transformer~\citep{zaheer2020big,choromanski2020rethinking} for certain tasks to allow more efficient learning. The details of the hyperparameters and model configurations are presented in Table~\ref{tab: lra-hyperparameters}. 
Note, the learning rate, number of epochs, batch size are all consistent with the default settings in LRA for both \textit{Big Birds}~\citep{zaheer2020big} and \textit{Performers}~\citep{choromanski2020rethinking}. For the rest of model configurations, see~\cite{tay2020long}.

\begin{table*}
    \centering
    \scalebox{0.65}{
    \begin{tabular}{l|ccc|ccc}
    \toprule
                       & \multicolumn{3}{c|}{\textit{Big Bird}} & \multicolumn{3}{c}{\textit{Performers}} \\
    \midrule
 Hyperparameters & Cifar 10 &  Path Finder 32 &  Text Classification & Cifar 10 &  Path Finder 32 &  Text Classification \\
    \midrule
 Learning Rate                  &   $5e^{-4}$   &   $5e^{-4}$   &   $2.5e^{-2}$  &  $5e^{-4}$   &  $3e^{-4}$&  $2.5e^{-2}$ \\
 \# of Epoch                    &   $200$       &   $200$       &   $625$        &  $200$       &  $200$  &  $625$\\
 Train \& Infer Batch Size      &   $256$       &   $512$       &   $32$         &  $256$       &  $512$  &  $32$\\
 \midrule
  \# of Layers                  & $1$           &   $4$         &   $1$          &  $1$         &  $4$    &  $1$ \\
  \# of Heads                   & $4$           &   $2$         &   $4$          &  $8$         &  $8$    &  $4$ \\
  Embedding Dimension           & $128$         &   $64$        &   $256$        &  $128$       &  $32$   &  $256$\\
  Query/Key/Value Dimension     & $64$          &   $32$        &   $256$        &  $64$        &  $16$   &  $256$\\
  Feedforward Network Dimension & $128$         &   $64$        &   $1024$       &  $128$       &  $32$   &  $1024$\\
  Block Size (specific to \textit{Big Birds})                    & $8$           &   $8$         &   $64$         &  --          &  --     &  --\\
\bottomrule
    
    \end{tabular}
    }
    \caption{LRA hyperparameters and model configurations for \textit{Big Bird} (left) and \textit{Performers} (right). Note. The learning rate, number of epoch, and the batch size are all consistent with default settings in LRA~\citep{tay2020long}. For the rest of model configurations, please see~\cite{tay2020long}.}
\label{tab: lra-hyperparameters}
    \label{tab:my_label}
\end{table*}



\section{Appendix: Experimental Results}\label{app:glue-experiments-results}

\begin{figure}[h]
\begin{center}
\includegraphics[scale=0.38]{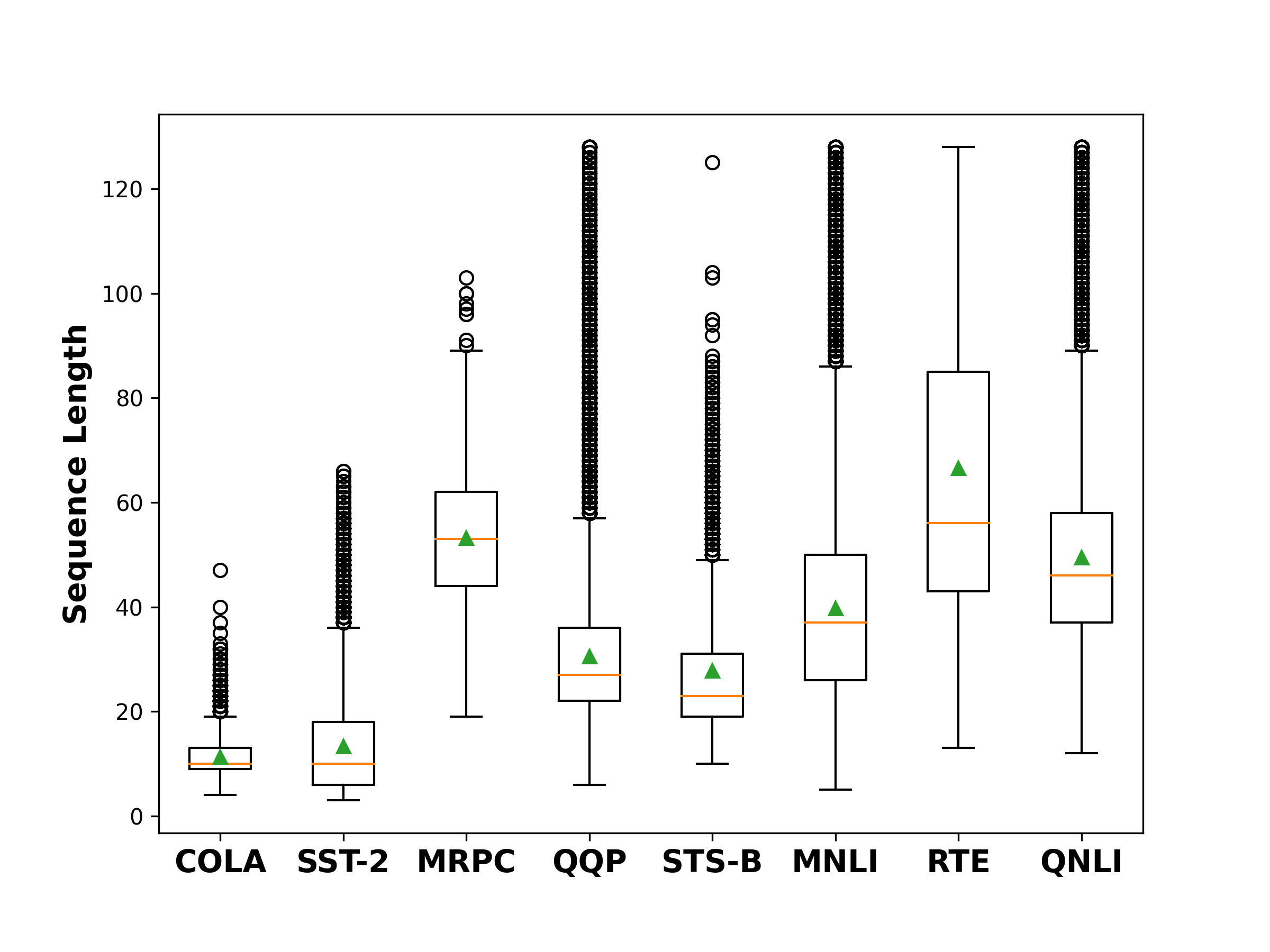}
\end{center}
\caption{Sequence length distributions of GLUE benchmarks. For each dataset, the triangle mark and solid line represent the mean and median, respectively.}
\label{fig: box-plot-seq-length-distribution}
\end{figure}

First, the GLUE \textit{dev} performance at $3.5X$, $3X$, $2X$, and $1.5X$ speedup are shown in Table~\ref{tab: glue-acc-speedup-3.5x},~\ref{tab: glue-acc-speedup-3x-appendix},~\ref{tab: glue-acc-speedup-2x}, and~\ref{tab: glue-acc-speedup-1.5x-appendix}. The same conclusion as discussed in Section~\ref{sec: results on glue datasets} is reached. Similarly, the GLUE \textit{dev} performance at $70\%$ and $30\%$ space complexity reduction are presented in the Table~\ref{tab: glue-acc-space-70-appendix} and~\ref{tab: glue-acc-space-30}. In particular, the baseline \textit{Average-pool} performs the worst in average across the GLUE benchmarks. This aligns with the proposed method from \citet{dai2020funnel} that a pre-training step is necessary to make the simple pooling method perform competitive. Especially for COLA dataset, \textit{Average-pool} shows the matthews correlation coefficients at most $25.6$ when the speedup is higher than $1.5X$. At $30\%$ space reduction in Table~\ref{tab: glue-acc-space-30}, the pooling method shows a relatively high matthews correlation coefficient of $44.6$, which is still the worst among all the other methods. We conjecture that the poor performance comes from the following reasons: (1) The pooling method, when applied at a encoder, significantly reduces the sequence length at least by half (as the least sliding window size is $2$). Additionally, the COLA dataset has the shortest sequence length in average in the GLUE benchmarks, making the reduced sequence length from the pooling even shorter. This fails the model to learn any useful pattern from the data. For the sequence length distributions of GLUE benchmarks, see Figure~\ref{fig: box-plot-seq-length-distribution}. (2) The reason the pooling method shows a relatively high matthews correlation coefficient at $30\%$ space reduction in Table~\ref{tab: glue-acc-space-30} is that the pruning scenario corresponds to a ``mild" one where the pooling is applied near the top encoder layer (after $8$th encoder) and the corresponded speedup is significantly less than $1.5X$. We also observed that the SST-2, which has the second shortest sequence length in average in the GLUE benchmarks, does not suffer similarly as the COLA for the pooling method. We conjecture that this is due to the task of sentiment analysis for SST-2 is much easier than that of judging the grammatical correctness of a sentence for COLA. In summary, the \textit{Average-pool} method merges the consecutive tokens together, according to the window length. However, in English text usually it is not the case that the consecutive tokens are similar and can be merged into one without significant loss in information.

\begin{figure*}[h]
\begin{center}
\includegraphics[width=1.0\textwidth, height=0.2\textheight]{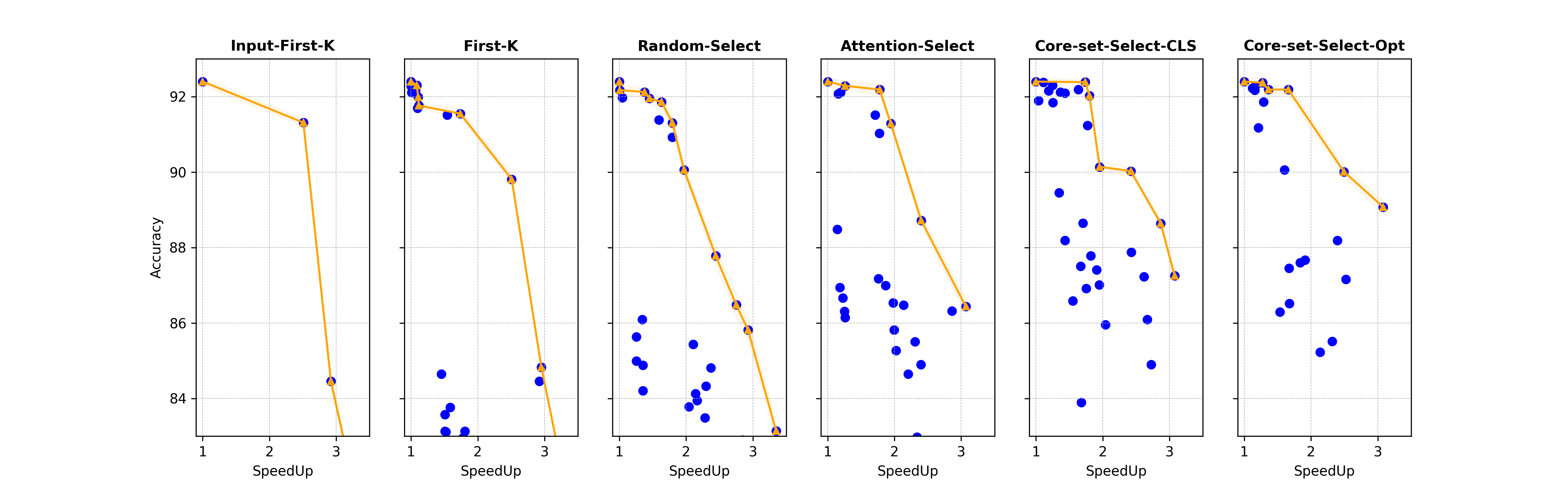}
\end{center}
\caption{SST-2: Pareto curves and scatter plots for accuracy vs. inference speedup trade-off. Each point corresponds to a sequence-length configuration in Table~\ref{tab: retention-config-specifics}. The pareto curves are used to generate accuracy vs. speedup Table~\ref{tab: glue-acc-speedup-3x},~\ref{tab: glue-acc-speedup-1.5x},~\ref{tab: glue-acc-speedup-3.5x},~\ref{tab: glue-acc-speedup-2x}. }
\label{fig:demo-scatter-pareto-sst-2}
\end{figure*}

\begin{figure*}[h]
\begin{center}
\includegraphics[width=1.0\textwidth, height=0.2\textheight]{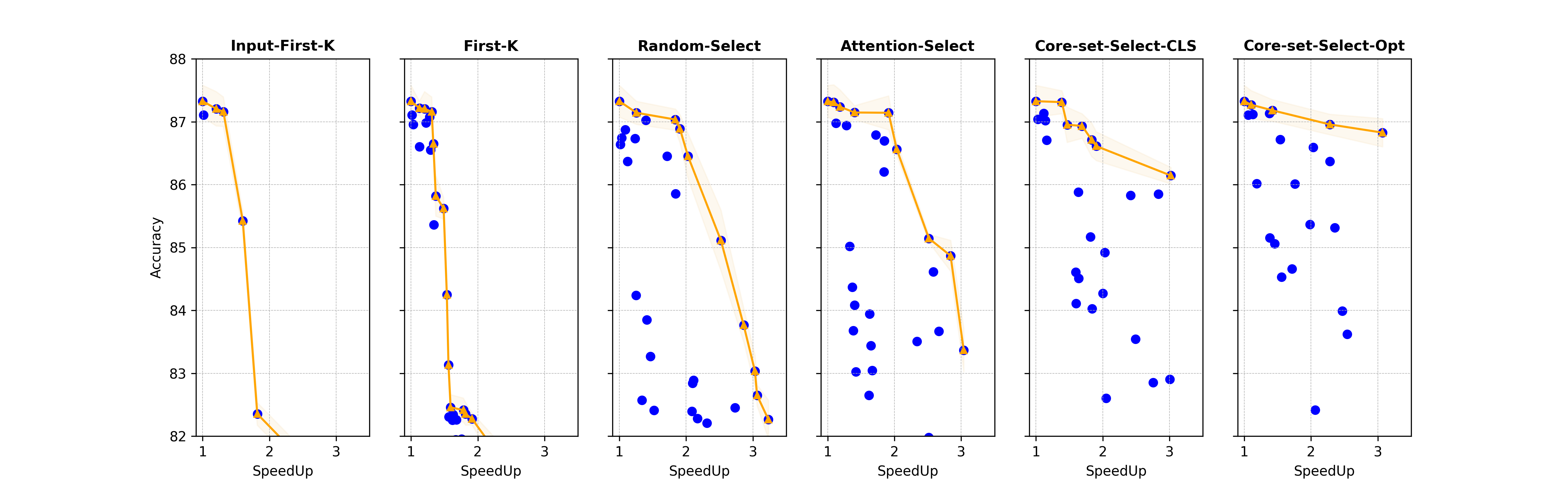}
\end{center}
\caption{MRPC: Pareto curves and scatter plots for accuracy vs. inference speedup trade-off. Each point corresponds to a sequence-length configuration in Table~\ref{tab: retention-config-specifics}. The pareto curves are used to generate accuracy vs. speedup Table~\ref{tab: glue-acc-speedup-3x},~\ref{tab: glue-acc-speedup-1.5x},~\ref{tab: glue-acc-speedup-3.5x},~\ref{tab: glue-acc-speedup-2x}.}
\label{fig:demo-scatter-pareto-mrpc}
\end{figure*}

Second, we demonstrate how to generate the accuracy and inference speedup table as shown in Table~\ref{tab: glue-acc-speedup-3x},~\ref{tab: glue-acc-speedup-1.5x},~\ref{tab: glue-acc-speedup-3.5x}, and~\ref{tab: glue-acc-speedup-2x}. A demonstration for dataset SST-2 and MRPC are presented in Figure~\ref{fig:demo-scatter-pareto-sst-2} and~\ref{fig:demo-scatter-pareto-mrpc}, respectively. Specifically, for each sequence-length configuration shown in Table~\ref{tab: retention-config-specifics}, we fine-tune a Transformer model with a token selection method on a GLUE \textit{train} set. Then we compute accuracy score and inference speedup number on its \textit{dev} set. A scatter plot of accuracy vs. speedup is made where each point corresponds to a sequence-length configuration. Next, among the scatter plot we find a pareto curve where accuracy scores at speedup $1.5X$, $2X$, $3X$, $3.5X$ are obtained. A linear interpolation is applied based on the pareto curve when necessary. However, we do not extrapolate the results for objective evaluation purpose. For \textit{Coreset-select-opt}, given a sequence-length configuration we choose the model that has the best accuracy score for $m\in\{\lceil0.1\cdot k\rceil, \lceil0.2\cdot k\rceil, \lceil0.3\cdot k\rceil, \lceil0.4\cdot k\rceil \}$. 

Third, for LRA, the results of accuracy scores for space complexity reduction at $70\%$ and $30\%$ are presented in Table~\ref{tab: long-text-70-reduction-all-appendix} and~\ref{tab: long-text-30-reduction-all}. At low space complexity reduction $30\%$, the advantages for \textit{Coreset-select-opt} over its baselines becomes smaller, which matches our expectations in the mild sequence-length reduction regime.


\begin{table*}
\centering
\scalebox{0.63}{
 \begin{tabular}{l|rrrrrrrrr|r}
\toprule
       METHOD &  STS-B &  MRPC &  SST-2 &  QNLI &  COLA &  RTE &  MNLI\_M &  MNLI\_MM &  QQP &  MEAN \\
\midrule
           \textit{Input-First-K-Select} &   $80.6 \pm 0.2$ &   -- &    -- &  $80.5 \pm 0.1$ &  $47.9 \pm 0.5$ &  -- &    $75.9 \pm 0.1$ &     $74.5 \pm 0.1$ &  -- &  $71.9$ \\
           \textit{First-K-Select} &   $80.3 \pm 0.2$ &   -- &    -- &  $80.8 \pm 0.1$ &  $47.9 \pm 0.5$ &  -- &    $75.5 \pm 0.1$ &     $74.7 \pm 0.1$ &  -- &  $71.8$ \\
            \textit{Random-Select} &   $85.7 \pm 0.2$ &   -- &    -- &  $86.2 \pm 0.1$ &  $48.9 \pm 0.5$ &  -- &    $80.2 \pm 0.1$ &     $78.1 \pm 0.1$ &  -- &  $75.8$ \\
            \textit{Average-Pool} &    $77.6 \pm 0.1$ & -- & -- & $84.0 \pm 0.1$ & $0.0 \pm 0.0$ & -- & $73.3 \pm 0.1$ & $74.7 \pm 0.0$ & -- & $61.9$ \\        
         \textit{Attention-Select} &   $85.3 \pm 0.1$ &   -- &    -- &  $86.7 \pm 0.1$ &  $51.0 \pm 0.4$ &  -- &    $81.1 \pm 0.1$ &     $81.4 \pm 0.1$ &  -- &  $77.1$ \\
 \textit{Coreset-Select-CLS (ours)} &   $\mathbf{86.6} \pm 0.1$ &   -- &    -- &  $\mathbf{87.5} \pm 0.1$ &  $51.1 \pm 0.3$ &  -- &    $\mathbf{82.2} \pm 0.1$ &    $\mathbf{82.4} \pm 0.1$ &  -- &  $77.9$ \\
\textit{Coreset-Select-Opt (ours)} &   $\mathbf{86.6} \pm 0.1$ &   -- &    -- &  $\mathbf{87.5} \pm 0.1$ &  $\mathbf{52.7} \pm 0.3$ &  -- &    $\mathbf{82.2} \pm 0.1$ &     $\mathbf{82.4} \pm 0.1$ &  -- &  $\mathbf{78.3}$ \\

 \midrule
              $\text{BERT}_\text{Base}$ &   $87.9 \pm 0.2$ &  -- &   -- &  $90.9 \pm 0.1$ &  $53.3 \pm 0.5$ &  -- &    $84.0 \pm 0.1$ &     $84.6 \pm 0.1$ & -- &  $80.2$ \\
\bottomrule
\end{tabular}
}
\caption{GLUE \textit{dev} performance at $3.5X$ speedup. Each score with standard deviation is averaged over $20$ trials. Larger score indicates better performance. The best score among each token selection method is embolden for each column. Note, missing values indicate that no accuracy score is observed for $3.5X$ speedup.}
\label{tab: glue-acc-speedup-3.5x}
\end{table*}

\begin{table*}
\centering
\scalebox{0.63}{
 \begin{tabular}{l|rrrrrrrrr|r}
\toprule
                                  METHOD &  STS-B &  MRPC &  SST-2 &  QNLI &  COLA &   RTE &  MNLI-M &  MNLI-MM &   QQP &  MEAN \\
\midrule
           \textit{Input-First-K-Select} &   $86.4 \pm 0.1$ &  $81.4 \pm 0.2$ &   $83.8 \pm 0.2$ &  $84.8 \pm 0.0$ &  $49.7  \pm 0.5$ &  $63.5 \pm 0.5$ & $77.8 \pm 0.1$ &  $75.9 \pm 0.0$ &  $80.8 \pm 0.1$ &  $76.0$ \\
           \textit{First-K-Select} &   $86.4 \pm 0.1$ &  $80.9 \pm 0.2$ &   $84.4 \pm 0.2$ &  $84.4 \pm 0.1$ &  $49.7 \pm 0.5$ &  $63.5 \pm 0.5$ &   $76.7 \pm 0.1$ &   $75.6 \pm 0.1$ &  $80.4 \pm 0.0$ &  $76.1$ \\
            \textit{Random-Select} &   $86.8 \pm 0.1$ &  $83.2 \pm 0.3$ &   $85.6 \pm 0.3$ &  $86.4 \pm 0.1$ &  $49.5 \pm 0.5$ & $62.1 \pm 0.8$ &   $81.4 \pm 0.1$ &  $78.7 \pm 0.1$ &  $87.0 \pm 0.1$ &  $77.9$ \\
            \textit{Average-Pool} &    $81.6 \pm 0.1$ & $83.9 \pm 0.3$ & $85.2 \pm 0.2$ & $84.1 \pm 0.1$ & $3.0 \pm 0.6$ & $59.2 \pm 0.5$ & $75.4 \pm 0.1$ & $76.7 \pm 0.1$ & $79.4 \pm 0.0$ & $69.6$ \\            
         \textit{Attention-Select} &   $\mathbf{87.0} \pm 0.1$ &  $84.6 \pm 0.3$ &   $86.0 \pm 0.2$ &  $86.8 \pm 0.1$ &  $51.1 \pm 0.4$ &  $63.4 \pm 0.6$ &   $\mathbf{82.5} \pm 0.1$ &     $\mathbf{82.7} \pm 0.1$ &  $\mathbf{87.3} \pm 0.1$ &  $79.0$ \\
 \textit{Coreset-Select-CLS (ours)} &   $\mathbf{87.0} \pm 0.1$ &  $86.2 \pm 0.1$ &   $87.3 \pm 0.1$ &  $\mathbf{87.8} \pm 0.1$ &  $51.7 \pm 0.3$ &  $\mathbf{63.7} \pm 0.5$ &   $82.4 \pm 0.1$ &    $82.6 \pm 0.1$ &  $\mathbf{87.3} \pm 0.0$ &  $79.6$ \\
 \textit{Coreset-Select-Opt (ours)} &   $\mathbf{87.0} \pm 0.1$ &  $\mathbf{86.9} \pm 0.2$ &   $\mathbf{89.6} \pm 0.1$ &  $\mathbf{87.8} \pm 0.1$ &  $\mathbf{52.8} \pm 0.3$ &  $\mathbf{63.7} \pm 0.5$ &    $\mathbf{82.5} \pm 0.1$ &     $\mathbf{82.7} \pm 0.1$ &  $\mathbf{87.3} \pm 0.0$ &  $\mathbf{80.0}$ \\
 \midrule
              $\text{BERT}_\text{Base}$ &   $87.9 \pm 0.2$ &  $87.3 \pm 0.2$ &   $92.4 \pm 0.1$ &  $90.9 \pm 0.1$ &  $53.3 \pm 0.5$ &  $65.8 \pm 0.5$ &    $84.0 \pm 0.1$ &     $84.6 \pm 0.1$ &  $87.5 \pm 0.1$ &  $81.5$ \\
\bottomrule
\end{tabular}
}
\caption{GLUE \textit{dev} performance at $3X$ speedup. Each score with standard deviation is averaged over $20$ trials. Larger score indicates better performance. The best score among each token selection method is embolden for each column.}
\label{tab: glue-acc-speedup-3x-appendix}
\end{table*}

\begin{table*}
\centering
\scalebox{0.63}{
 \begin{tabular}{l|rrrrrrrrr|r}
\toprule
        METHOD &  STS-B &  MRPC &  SST-2 &  QNLI &  COLA &   RTE &  MNLI-M &  MNLI-MM &   QQP &  MEAN \\
\midrule
           \textit{Input-First-K-Select} &   $87.8 \pm 0.2$ &  $82.2 \pm 0.2$ &   $91.8 \pm 0.1$ &  $\mathbf{90.4} \pm 0.1$ &  $51.9 \pm 0.5$ &  $65.2 \pm 0.5$ &    $81.9 \pm 0.1$ &     $82.2 \pm 0.1$ &  $85.4 \pm 0.1$ &  $79.9$ \\
           \textit{First-K-Select} &   $87.8\pm 0.2$ &  $82.1 \pm 0.2$ &   $91.2 \pm 0.1$ &  $\mathbf{90.4} \pm 0.1$ &  $51.9 \pm 0.5$ &  $64.9 \pm 0.5$ &    $81.9 \pm 0.1$ &     $82.2 \pm 0.1$ &  $84.9 \pm 0.0$ &  $79.7$ \\
            \textit{Random-Select} &   $87.8\pm 0.1$ &  $86.6 \pm 0.3$ &   $90.0 \pm 0.2$ &  $88.9 \pm 0.1$ &  $52.6 \pm 0.5$ &  $64.9 \pm 0.5$ &    $83.6 \pm 0.1$ &     $80.3 \pm 0.1$ &  $87.2 \pm 0.1$ &  $80.2$ \\
            \textit{Average-Pool} &    $87.8 \pm 0.2$ & $85.8 \pm 0.3$ & $88.6 \pm 0.1$ & $86.3 \pm 0.1$ & $11.2 \pm 0.7$ & $60.2 \pm 0.6$ & $76.3 \pm 0.0$ & $82.9 \pm 0.0$ & $82.1 \pm 0.0$ & $73.4$ \\            
         \textit{Attention-Select} &   $87.8\pm 0.1$ &  $86.8 \pm 0.2$ &   $91.0 \pm 0.1$ &  $90.4 \pm 0.1$ &  $\mathbf{53.2} \pm 0.5$ &  $\mathbf{65.7} \pm 0.6$ &    $83.6 \pm 0.1$ &     $84.1 \pm 0.1$ &  $87.3 \pm 0.1$ &  $81.1$ \\
 \textit{Coreset-Select-CLS (ours)} &   $87.6\pm 0.1$ &  $86.6 \pm 0.2$ &   $90.1 \pm 0.1$ &  $89.4 \pm 0.1$ &  $\mathbf{53.2} \pm 0.5$ &  $64.1 \pm 0.5$ &    $\mathbf{83.9} \pm 0.1$ &    $\mathbf{84.3} \pm 0.1$ &  $\mathbf{87.5} \pm 0.1$ &  $80.7$ \\
 \textit{Coreset-Select-Opt (ours)} &   $\mathbf{87.9}\pm 0.1$ &  $\mathbf{87.2} \pm 0.1$ &   $\mathbf{92.4} \pm 0.1$ &  $89.4 \pm 0.1$ &  $\mathbf{53.2} \pm 0.5$ &  $64.9 \pm 0.5$ &   $\mathbf{83.9} \pm 0.1$ &     $\mathbf{84.3} \pm 0.1$ &  $\mathbf{87.5} \pm 0.0$ &  $\mathbf{81.2}$ \\
 \midrule
              $\text{BERT}_\text{Base}$ &   $87.9 \pm 0.2$ &  $87.3 \pm 0.2$ &   $92.4 \pm 0.1$ &  $90.9 \pm 0.1$ &  $53.3 \pm 0.5$ &  $65.8 \pm 0.5$ &    $84.0 \pm 0.1$ &     $84.6 \pm 0.1$ &  $87.5 \pm 0.1$ &  $81.5$ \\
\bottomrule
\end{tabular}
}
\caption{GLUE \textit{dev} performance at $2X$ speedup. Each score with standard deviation is averaged over $20$ trials. Larger score indicates better performance. The best score among each token selection method is embolden for each column.}
\label{tab: glue-acc-speedup-2x}
\end{table*}

\begin{table*}
\centering
\scalebox{0.63}{
 \begin{tabular}{l|rrrrrrrrr|r}
\toprule
        METHOD &  STS-B &  MRPC &  SST-2 &  QNLI &  COLA &   RTE &  MNLI-M &  MNLI-MM &   QQP &  MEAN \\
\midrule
           \textit{Input-First-K-Select} &   $\mathbf{87.9} \pm 0.2$ &  $86.8 \pm 0.2$ &   $92.1 \pm 0.1$ &  $90.8 \pm 0.1$ &  $53.0 \pm 0.5$ &  $65.6 \pm 0.5$ &    $\mathbf{84.0} \pm 0.1$ &     $84.1 \pm 0.1$ &  $87.1 \pm 0.1$ &  $81.3$ \\
           \textit{First-K-Select} &   $\mathbf{87.9} \pm 0.2$ &  $86.4 \pm 0.2$ &   $91.5 \pm 0.1$ &  $90.8 \pm 0.1$ &  $52.7 \pm 0.4$ &  $65.2 \pm 0.5$ &    $83.8 \pm 0.1$ &     $84.0 \pm 0.1$ &  $86.9 \pm 0.0$ &  $81.0$ \\
            \textit{Random-Select} &   $87.8 \pm 0.1$ &  $87.2 \pm 0.2$ &   $91.9 \pm 0.2$ &  $90.8 \pm 0.1$ &  $53.1 \pm 0.4$ &  $\mathbf{65.7} \pm 0.5$ &    $83.9 \pm 0.1$ &     $83.9 \pm 0.1$ &  $87.4 \pm 0.1$ &  $81.3$ \\
            \textit{Average-Pool} &    $87.8 \pm 0.1$ & $87.0 \pm 0.2$ & $90.3 \pm 0.1$ & $90.2 \pm 0.1$ & $25.6 \pm 0.7$ & $61.5 \pm 0.6$ & $80.9 \pm 0.1$ & $84.0 \pm 0.0$ & $85.7 \pm 0.0$ & $76.8$ \\            
         \textit{Attention-Select} &   $\mathbf{87.9} \pm 0.1$ &  $87.1 \pm 0.1$ &   $92.3 \pm 0.1$ &  $90.7 \pm 0.1$ &  $53.2 \pm 0.5$ &  $\mathbf{65.7} \pm 0.5$ &    $\mathbf{84.0} \pm 0.1$ &     $84.5 \pm 0.1$ & $87.4 \pm 0.1$ &  $81.4$ \\
 \textit{Coreset-Select-CLS (ours)} &   $87.7 \pm 0.1$ &  $86.9 \pm 0.2$ &   $\mathbf{92.4} \pm 0.1$ &  $\mathbf{90.9} \pm 0.1$ &  $\mathbf{53.3} \pm 0.5$ &  $65.4 \pm 0.3$ & $\mathbf{84.0} \pm 0.0$ &     $\mathbf{84.6} \pm 0.1$ &  $\mathbf{87.5} \pm 0.1$ &  $81.4$ \\
 \textit{Coreset-Select-Opt (ours)} &   $\mathbf{87.9} \pm 0.1$ &  $\mathbf{87.3} \pm 0.2$ &   $\mathbf{92.4} \pm 0.1$ &  $\mathbf{90.9} \pm 0.1$ &  $\mathbf{53.3} \pm 0.5$ &  $65.6 \pm 0.5$ &    $\mathbf{84.0} \pm 0.0$ &    $\mathbf{84.6} \pm 0.1$ &  $\mathbf{87.5} \pm 0.0$ &  $\mathbf{81.5}$ \\
 \midrule
              $\text{BERT}_\text{Base}$ &   $87.9 \pm 0.2$ &  $87.3 \pm 0.2$ &   $92.4 \pm 0.1$ &  $90.9 \pm 0.1$ &  $53.3 \pm 0.5$ &  $65.8 \pm 0.5$ &    $84.0 \pm 0.1$ &     $84.6 \pm 0.1$ &  $87.5 \pm 0.1$ &  $81.5$ \\
\bottomrule
\end{tabular}
}
\caption{GLUE \textit{dev} performance at $1.5X$ speedup. Each score with standard deviation is averaged over $20$ trials. Larger score indicates better performance. The best score among each token selection method is embolden for each column.}
\label{tab: glue-acc-speedup-1.5x-appendix}
\end{table*}

\begin{table*}
\centering
\scalebox{0.63}{
 \begin{tabular}{l|rrrrrrrrr|r}
\toprule
        METHOD &  STS-B &  MRPC &  SST-2 &  QNLI &  COLA &   RTE &  MNLI-M &  MNLI-MM &   QQP &  MEAN \\
\midrule
           \textit{Input-first-k-select} &   $85.3 \pm 0.1$ &  $81.3 \pm 0.2$ &   $83.3 \pm 0.2$ &  $84.6 \pm 0.0$ &  $49.0 \pm 0.4$ &  $62.1 \pm 0.5$ & $76.9 \pm 0.1$ &   $74.9 \pm 0.1$ &  $80.6 \pm 0.1$ &  $75.3$ \\
        \textit{First-k-select}   &   $85.1 \pm 0.1$ &  $81.5 \pm 0.2$ &   $84.6 \pm 0.2$ &  $84.3 \pm 0.1$ &  $49.0 \pm 0.4$ &  $62.0 \pm 0.5$ &   $76.3 \pm 0.1$ &  $74.5 \pm 0.1$ &  $80.0 \pm 0.0$ &  $75.3$ \\
            \textit{Random-select} &   $85.6 \pm 0.2$ &  $83.3 \pm 0.3$ &   $84.9 \pm 0.2$ &  $85.1 \pm 0.1$ &  $48.4 \pm 0.4$ &  $61.8 \pm 0.5$ &  $79.0 \pm 0.1$ & $79.3 \pm 0.1$ &  $86.6 \pm 0.1$ &  $77.1$ \\
         \textit{Average-Pool} &    $78.7 \pm 0.1$ & $83.1 \pm 0.2$ & $85.1 \pm 0.2$ & $84.0 \pm 0.1$ & $0.0 \pm 0.0$ & $59.8 \pm 0.6$ & $75.2 \pm 0.1$ & $76.3 \pm 0.1$ & $82.9 \pm 0.1$ & $69.5$ \\  
         \textit{Attention-select} &   $85.4 \pm 0.1$ &  $84.3 \pm 0.2$ &   $87.2 \pm 0.1$ &  $86.4 \pm 0.1$ &  $50.9 \pm 0.4$ &  $62.7 \pm 0.4$ &    $80.5 \pm 0.1$ &    $80.7 \pm 0.1$ &  $87.0 \pm 0.1$ &  $78.3$ \\
 \textit{Coreset-Select-CLS (ours)} &   $86.5 \pm 0.1$ &  $86.0 \pm 0.1$ &   $87.6 \pm 0.1$ &  $86.6 \pm 0.1$ &  $51.0 \pm 0.4$ &  $\mathbf{63.6} \pm 0.4$ &    $80.9 \pm 0.1$ & $81.6 \pm 0.1$ &  $87.2 \pm 0.1$ &  $79.0$ \\
 \textit{Coreset-Select-Opt (ours)} &   $\mathbf{86.7} \pm 0.1$ &  $\mathbf{86.6} \pm 0.2$ &   $\mathbf{87.7} \pm 0.1$ &  $\mathbf{86.5} \pm 0.1$ &  $\mathbf{52.3} \pm 0.4$ & $\mathbf{63.6} \pm 0.4$ &    $\mathbf{81.0} \pm 0.1$ &     $\mathbf{81.8} \pm 0.1$ &  $\mathbf{87.3} \pm 0.0$ &  $\mathbf{79.3}$ \\
 \midrule
              $\text{BERT}_\text{Base}$ &   $87.9 \pm 0.2$ &  $87.3 \pm 0.2$ &   $92.4 \pm 0.1$ &  $90.9 \pm 0.1$ &  $53.3 \pm 0.5$ &  $65.8 \pm 0.5$ &    $84.0 \pm 0.1$ &     $84.6 \pm 0.1$ &  $87.5 \pm 0.1$ &  $81.5$ \\
\bottomrule
\end{tabular}
}
\caption{GLUE \textit{dev} performance at at $70 \%$ space complexity reduction. Each score with standard deviation is averaged over $20$ trials. Larger score indicates better performance. The best score among each token selection method is embolden for each column.}
\label{tab: glue-acc-space-70-appendix}
\end{table*}

\begin{table*}
\centering
\scalebox{0.63}{
 \begin{tabular}{l|rrrrrrrrr|r}
\toprule
        METHOD &  STS-B &  MRPC &  SST-2 &  QNLI &  COLA &   RTE &  MNLI-M &  MNLI-MM &   QQP &  MEAN \\
\midrule
           \textit{Input-first-k-select} &   $\mathbf{87.9} \pm 0.2$ &  $87.2 \pm 0.2$ &   $91.8 \pm 0.1$ &  $90.8 \pm 0.1$ &  $53.0 \pm 0.5$ &  $65.2 \pm 0.5$ &    $\mathbf{84.0} \pm 0.1$ &     $\mathbf{84.3} \pm 0.1$ &  $87.1 \pm 0.2$ &  $81.3$ \\
        \textit{First-k-select}   &   $\mathbf{87.9} \pm 0.2$ &  $87.2 \pm 0.1$ &   $91.3 \pm 0.1$ &  $90.9 \pm 0.1$ &  $53.0 \pm 0.5$ &  $65.2 \pm 0.5$ &    $\mathbf{84.0} \pm 0.1$ &    $\mathbf{84.3} \pm 0.1$ &  $87.4 \pm 0.1$ &  $81.2$ \\
            \textit{Random-select} &   $87.8 \pm 0.1$ &  $87.1 \pm 0.2$ &   $92.1 \pm 0.1$ &  $\mathbf{90.9} \pm 0.2$ &  $53.1 \pm 0.4$ &  $\mathbf{65.7} \pm 0.5$ &    $83.9 \pm 0.1$ &     $84.4 \pm 0.1$ &  $87.4 \pm 0.2$ &  $81.4$ \\
         \textit{Average-Pool} &    $87.8 \pm 0.1$ & $87.2 \pm 0.1$ & $91.7 \pm 0.1$ & $89.2 \pm 0.1$ & $44.6 \pm 0.5$ & $64.1 \pm 0.5$ & $82.7 \pm 0.1$ & $83.7 \pm 0.1$ & $86.2 \pm 0.1$ & $79.7$ \\  
         \textit{Attention-select} &   $\mathbf{87.9} \pm 0.1$ &  $87.1 \pm 0.1$ &   $92.3 \pm 0.1$ &  $90.7 \pm 0.1$ &  $\mathbf{53.2} \pm 0.4$ &  $\mathbf{65.7} \pm 0.5$ &    $83.9 \pm 0.1$ &     $84.4 \pm 0.1$ &  $\mathbf{87.5} \pm 0.2$ &  $81.4$ \\
 \textit{Coreset-Select-CLS (ours)} &   $87.7 \pm 0.1$ &  $\mathbf{87.3} \pm 0.2$ &   $\mathbf{92.4} \pm 0.1$ &  $\mathbf{90.9} \pm 0.1$ &  $\mathbf{53.2} \pm 0.4$ &  $65.5 \pm 0.3$ &    $\mathbf{84.0} \pm 0.1$ &     $\mathbf{84.3} \pm 0.1$ &  $\mathbf{87.5} \pm 0.1$ &  $81.4$ \\
 \textit{Coreset-Select-Opt (ours)} &   $\mathbf{87.9} \pm 0.1$ &  $\mathbf{87.3} \pm 0.2$ &   $\mathbf{92.4} \pm 0.1$ &  $\mathbf{90.9} \pm 0.1$ &  $\mathbf{53.2} \pm 0.4$ &  $65.6 \pm 0.5$ &    $\mathbf{84.0} \pm 0.1$ &     $\mathbf{84.3} \pm 0.1$ &  $\mathbf{87.5} \pm 0.1$ &  $\mathbf{81.5}$ \\
 \midrule
              $\text{BERT}_\text{Base}$ &   $87.9 \pm 0.2$ &  $87.3 \pm 0.2$ &   $92.4 \pm 0.1$ &  $90.9 \pm 0.1$ &  $53.3 \pm 0.5$ &  $65.8 \pm 0.5$ &    $84.0 \pm 0.1$ &     $84.6 \pm 0.1$ &  $87.5 \pm 0.1$ &  $81.5$ \\
\bottomrule
\end{tabular}
}
\caption{GLUE \textit{dev} performance at at $30 \%$ space complexity reduction. Each score with standard deviation is averaged over $20$ trials. Larger score indicates better performance. The best score among each token selection method is embolden for each column.}
\label{tab: glue-acc-space-30}
\end{table*}

\begin{table*}
    \centering
    \scalebox{0.55}{
    \begin{tabular}{l|ccc|c|ccc|c}
    \toprule
                       & \multicolumn{4}{c|}{\textit{BIG BIRD}} & \multicolumn{4}{c}{\textit{PERFORMERS}} \\
    \midrule
 METHOD & CIFAR-10 &  PATHFINDER-32 &  IMDB (BYTE-LEVEL) &  MEAN &  CIFAR-10 &  PATHFINDER-32 & IMDB (BYTE-LEVEL) &  MEAN \\
    \midrule
\textit{First-k-select} &       $26.9 \pm 0.2$ &    $55.6  \pm 0.2$ &    $57.9 \pm 0.1$ & $46.8$ &$26.9 \pm 0.1$ & $52.4 \pm 0.2$ & $59.9 \pm 0.1$ & $46.4$  \\
    
\textit{Random-select} &      $39.4 \pm 0.1$ &  $69.9 \pm 0.2$ & $59.6 \pm 0.1$ &  $56.3$  & $41.5 \pm 0.2$ &   $58.2 \pm 0.1$ & $59.9 \pm 0.1$ &  $53.2$ \\
    
\textit{Coreset-Select-CLS (ours)} &     $38.6 \pm 0.1$ &   $69.3 \pm 0.2$ &    $59.1 \pm 0.1$ &  $55.7$   & $39.8 \pm 0.1$ &   $61.5 \pm 0.2$ &    $59.7 \pm 0.1$ &  $53.7$\\
    
 \textit{Coreset-Select-Opt (ours)} &   $\mathbf{43.3} \pm 0.2$ &   $\mathbf{71.7} \pm 0.1$ &   $\mathbf{61.4} \pm 0.2$ &  $\mathbf{58.8}$   &  $\mathbf{45.5} \pm 0.1$ &   $\mathbf{67.7} \pm 0.1$ &                 $\mathbf{62.8} \pm 0.1$ &  $\mathbf{58.7}$ \\
    \midrule
$\text{Trans-No-Prune}$    &      $40.9 \pm 0.1$ &          $73.5 \pm 0.2$ &        $63.8 \pm 0.1$ &  $59.4$   &     $42.9 \pm 0.1$ &            $66.2 \pm 0.2$ &                 $64.3 \pm 0.1$ &  $57.8$ \\
\bottomrule
    
    \end{tabular}
    }
    \caption{LRA \textit{test} set performances at $70 \%$ space complexity reduction for \textit{Big Bird} (left) and \textit{Performers} (right) as the backbone Transformer. Each score with standard deviation is averaged over $20$ trials. Larger score indicates better performance. The best score among each token selection method is embolden for each column.}
\label{tab: long-text-70-reduction-all-appendix}
    \label{tab:my_label}
\end{table*}

\begin{table*}
    \centering
    \scalebox{0.55}{
    \begin{tabular}{l|ccc|c|ccc|c}
    \toprule
                       & \multicolumn{4}{c|}{\textit{BIG BIRD}} & \multicolumn{4}{c}{\textit{PERFORMERS}} \\
    \midrule
 METHOD & CIFAR-10 &  PATHFINDER-32 &  IMDB (BYTE-LEVEL) &  MEAN &  CIFAR-10 &  PATHFINDER-32 & IMDB (BYTE-LEVEL) &  MEAN \\
    \midrule
\textit{First-k-select} &       $37.2 \pm 0.1$ &            $69.7 \pm 0.2$ &    $\mathbf{62.8} \pm 0.1$ &  $56.6$ &      $37.6 \pm 0.1$ &   $64.6 \pm 0.2$ & $63.0 \pm 0.1$ &  $55.1$  \\
    
\textit{Random-select} &      $41.1 \pm 0.2$ &            $70.0 \pm 0.2$ &    $62.5 \pm 0.1$ &  $57.9$  &      $43.2 \pm 0.1$ &     $65.6 \pm 0.2$ &    $63.3 \pm 0.1$ &  $57.4$ \\
    
\textit{Coreset-Select-CLS (ours)} &     $40.4 \pm 0.2$ &            $\mathbf{72.3} \pm 0.2$ &  $62.2 \pm 0.1$ &  $58.3$   & $41.9 \pm 0.2$ &  $66.6 \pm 0.1$ &    $63.1 \pm 0.1$ &  $57.2$\\
    
 \textit{Coreset-Select-Opt (ours)} &       $\mathbf{43.7} \pm 0.1$ &            $\mathbf{72.3} \pm 0.2$ &                 $62.7 \pm 0.1$ &  $\mathbf{59.6}$   &       $\mathbf{46.1} \pm 0.1$ &    $\mathbf{68.5} \pm 0.2$ &   $\mathbf{63.9} \pm 0.1$ &  $\mathbf{59.5}$ \\
    \midrule
$\text{Trans-No-Prune}$    &      $40.9 \pm 0.1$ &          $73.5 \pm 0.2$ &        $63.8 \pm 0.1$ &  $59.4$   &     $42.9 \pm 0.1$ &            $66.2 \pm 0.2$ &                 $64.3 \pm 0.1$ &  $57.8$ \\
\bottomrule
    
    \end{tabular}
    }
    \caption{LRA \textit{test} set performances at $30 \%$ space complexity reduction for \textit{Big Bird} (left) and \textit{Performers} (right) as the backbone Transformer. Each score with standard deviation is averaged over $20$ trials. Larger score indicates better performance. The best score among each token selection method is embolden for each column.}
\label{tab: long-text-30-reduction-all}
    \label{tab:my_label}
\end{table*}

\section{Appendix: Experimental Details for Figure~\ref{fig:cls_similarity} in Section~\ref{sec:our model} }\label{app:explain-figure-1}

A preliminary study of token embeddings show that as the embeddings propagate through the pipeline of encoders, they become more and more similar to the CLS token, Figure  \ref{fig:cls_similarity} (top row). A deeper investigation shows that they also become more and more similar with each other and form clusters among themselves, Figure \ref{fig:cls_similarity} (bottom row). 

To obtain Figure~\ref{fig:cls_similarity}, we examine token embeddings of the SST-2 \textit{dev} set after fine-tuning a $\text{BERT}_{\text{Base}}$ on its \textit{train} set. For Figure~\ref{fig:cls_similarity} (top row), we compute the histogram of cosine similarity between the embedding of CLS and that of all the other tokens in encoder $1$, $6$, and $12$, respectively. For Figure~\ref{fig:cls_similarity} (bottom row), we apply DBSCAN~\cite{ester1996density} to cluster the embeddings of tokens in encoder $1$, $6$, and $12$ respectively. For the hyper-parameters of DBSCAN, the maximum distance $\epsilon$ of two points in a cluster, the minimum number of points required to form a cluster are set as $0.2$ and $1$. The distance metric is cosine dissimilarity.

\end{document}